\documentclass[11pt]{scrartcl}       
\usepackage[bottom=2.5cm,top=2.5cm,left=2.5cm,right=2.5cm]{geometry}

\newcommand{\twohalf}{.3}

\usepackage{amsmath,amssymb,amsthm,graphicx,caption,cases}	
\usepackage[bottom]{footmisc}	
\usepackage{algorithmic}

\usepackage{abstract,amsfonts,amsbsy,amssymb,amstext,mathrsfs,latexsym,color,url,amsthm,xcolor,authblk,fontenc,bm}

\definecolor{NUSBlue}{RGB}{0,61,124} 
\definecolor{NUSOrange}{RGB}{239,124,0}
\usepackage[colorlinks=false,allbordercolors=NUSOrange]{hyperref}

\DeclareOldFontCommand{\bf}{\normalfont\bfseries}{\mathbf}

\DeclareMathOperator{\diag}{diag}

\newcommand{\bI}{\mathbf{I}}

\newcommand{\bD}{\mathbf{D}}
\newcommand{\bL}{\mathbf{L}}

\newcommand{\bW}{\mathbf{W}}

\newcommand{\bb}{\mathbf{b}}

\newcommand{\bA}{\mathbf{A}}

\newcommand{\bx}{\mathbf{x}}
\newcommand{\by}{\mathbf{y}}

\newcommand{\cK}{\mathcal{K}}

\newcommand{\cB}{\mathcal{B}}

\newcommand{\R}{\mathbb R}

\definecolor{NUSBlue}{RGB}{0,61,124}   

\usepackage{tikz}
\usepackage{mathtools}
\usepackage{color}

\def\nudge{.5}

\tikzset{axis/.style={ultra thin, Grey, -latex, shorten <=-\nudge cm, shorten >=-2*\nudge cm}}
\tikzset{line/.style={thick}}

\usepackage{textcomp}
\DeclareMathAlphabet\mathbfcal{OMS}{cmsy}{b}{n}
\usepackage{hyperref}
\usepackage[linesnumbered,vlined,ruled]{algorithm2e}
\usepackage{amsmath,amsfonts,amssymb,amsthm}
\theoremstyle{plain}
\newtheorem{thm}{Theorem}

\newtheorem{lem}[thm]{Lemma}

\newtheorem{exam}[thm]{Example}

\newtheoremstyle{cited}%
  {3pt}
  {3pt}
{\itshape}
  {}
  {\bfseries}
  {.}
  {.5em}
  {\thmname{#1} \thmnumber{#2} \thmnote{\normalfont#3}}
\theoremstyle{cited}
\newtheorem{citedthm}[thm]{Theorem}
\newtheorem{citedlem}[thm]{Lemma}

\newtheorem{citeddef}[thm]{Definition}

\usepackage{graphicx}
\usepackage{epstopdf}
\usepackage{epsf}
\usepackage{subfigure}
\usepackage{epsfig}

\usepackage{cite}

\usepackage{tabularx}
\usepackage{array}
\usepackage{booktabs}

\usepackage{comment}
\usepackage{tgpagella} 
\usepackage{mathpazo}

\begin{document}
\renewcommand*{\Authsep}{, }
\renewcommand*{\Authand}{, }
\renewcommand*{\Authands}{, }
\renewcommand*{\Affilfont}{\normalsize}   
\setlength{\affilsep}{2em}   
\title{Analysis of function approximation and stability of general DNNs in directed acyclic graphs using un-rectifying analysis}
\date{\today}
\author{Wen-Liang Hwang and Shih-Shuo Tung}
\maketitle

\begin{abstract}
A general lack of understanding pertaining to deep feedforward neural networks (DNNs) can be attributed partly to a lack of tools with which to analyze the composition of non-linear functions, and partly to a lack of mathematical models applicable to the diversity of DNN architectures. In this paper, we made a number of basic assumptions pertaining to activation functions, non-linear transformations, and DNN architectures in order to use the un-rectifying method to analyze DNNs via directed acyclic graphs (DAGs). DNNs that satisfy these assumptions are referred to as general DNNs. Our construction of an analytic graph was based on an axiomatic method in which DAGs are built from the bottom-up through the application of atomic operations to basic elements in accordance with regulatory rules. This approach allows us to derive the properties of general DNNs via mathematical induction.
We show that using the proposed approach, some properties hold true for general DNNs can be derived. 
This analysis advances our understanding of network functions and could promote further theoretical insights if the host of analytical tools for graphs can be leveraged.
\end{abstract}

%
%
%

\section{Introduction}

Deep feedforward neural networks (DNNs) have revolutionized the use of machine learning in many fields, such as computer vision, and signal processing, where they have been used to resolve ill-posed inverse problems and sparse recovery problems \cite{sun2016deep,monga2021algorithm}. However, researchers have yet to elucidate several fundamental issues that are critical to the way that DNNs function. 
This lack of understanding can be attributed at least partially to a lack of tools by which to analyze the composition of non-linear activation functions in DNNs, and a lack of mathematical models applicable to the diversity of DNN architectures. This paper reports on a preliminary study of fundamental issues pertaining to function approximation and the stability inherent of DNNs.

Simple series-connected DNN models, such as $ \mathcal N^s_L(\bx) 
= \varrho_L \circ M_L \circ\cdots\circ\varrho_1\circ M_1(\bx)
$, 
are widely adopted for analysis \cite{BaraniukPowerDiagramSubdiv, Wen19,li2019deep}.
We initially considered whether the theoretical results derived in those papers were intrinsic/common to all DNNs or unique to this type of network. However, our exploration of this issue was hindered by problems encountered in representing a large class of DNN architectures and formulating the computation spaces for activation functions and non-linear transformations. In the current work, we addressed the latter problem by assuming that all activation functions can be expressed as networks with point-wise CPWL activation functions, while assuming that all non-linear transforms are Lipschitz functions. The difficulties involved in covering all possible DNNs prompted us to address the former problem by associating DNNs with graphs that can be described in a bottom-up manner, using an axiomatic approach, thereby allowing analysis of each step in the construction process. This approach made it possible for us to build complex networks from simple ones and derive their intrinsic properties using mathematical induction.

We sought to avoid generating graphs with loops by describing DNNs using directed acyclic graphs (DAGs).
The arcs are associated with basic elements that correspond to operations applied to the layers of a DNN (e.g., linear matrix, affine linear matrix, non-linear activation function/transformation) and nodes that delineate basic elements are used to relay and reshape the dimension of an input or combine outputs from incoming arcs to outgoing arcs. We refer to DNNs that can be constructed using the proposed axiomatic approach as general DNNs. It is unclear whether general DNNs are equivalent to all DNNs that are expressible using DAGs. Nevertheless, general DNNs include modules widely employed in well-known DNN architectures. The proposed approach makes it possible to extend the theoretical results for series-connected DNNs to general DNNs, as follows:
\begin{itemize}
\item  A DNN DAG divides the input space via partition refinement using either a composition of activation functions along a path or a fusion operation combining inputs from more than one path in the graph.  This makes it possible to approximate a target function in a coarse-to-fine manner by applying a local approximating function to each partition of the input space. 
\item Under mild assumptions related to point-wise CPWL activation functions and non-linear transformations, the stability of a DNN against local input perturbations can be maintained using sparse/compressible weight coefficients associated with incident arcs to a node. 
\end{itemize}

Taken together, we can conclude that a general DNN ``divides" the input space, ``conquers" the target function by applying a simple approximating function over each partition region, and ``sparsifies" weight coefficients to ensure robustness against input perturbations.

In the literature, graphs are commonly used to elucidate the structure of DNNs; however, they are seldom used to further analysis of DNNs. It should be noted that graph DNNs \cite{kipf2016semi} and our approach both adopt graphs for analysis;
however, graph DNNs focus on the operations of neural networks in order to represent real-world datasets in graphs (e.g., social networks and molecular structure \cite{duvenaud2015convolutional}), whereas our approach focuses on the construction of analyzable graph representations by which to deduce intrinsic properties of DNNs. 
In contrast, spectral graph theory \cite{defferrard2016convolutional} provides the mathematical background and numerical methods required for the design of fast localized convolutional filters for the data in graphs.

The remainder of the paper is organized as follows.  
In Section~\ref{sec:rw}, we review related works. Section~\ref{DAGsec} presents our bottom-up axiomatic approach to the construction of DNNs. Section~\ref{commonprop} outlines the function approximation and stability of general DNNs. Concluding remarks are presented in Section \ref{conclusions}.

\textbf{Notation:} \\
Matrices are denoted using bold upper case and vectors are denoted using bold lower case. We also use $x_i$ to denote the $i$-th entry of a vector $\bx\in\R^n$, $\|\bx\|_2$ to denote its Euclidean norm, and $\diag(\bx)$ to denote the diagonal matrix with diagonal $\bx$. \\

\section{Related works} \label{sec:rw}

Below, we review analytical methods applicable to network design, deriving properties, and achieving a more comprehensive understanding of DNNs.



The ordinary differential equation (ODE) approach was originally inspired by the residual network (ResNet)\cite{he2016deep}, wherein the latter is regarded as a discrete implementation of an ODE \cite{weinan2017proposal}. 
The ODE approach can be used to interpret networks by regarding them as different discretizations of different ODEs. 
Note that by designing numerical methods for ODEs, it is possible to develop new network architecture \cite{lu2018beyond}. The tight connection between the ODE and dynamic systems \cite{haber2017stable} makes it possible to study the stability of forward inference in a DNN and the well-posedness of learning a DNN (i.e., whether a DNN can be generalized by adding appropriate regularizations or training data), in which the stability of the DNN is related to initial conditions and network design is related to the design of a system of the ODE system. The ODE approach can also be used to study recurrent networks  \cite{chang2019antisymmetricrnn}. Nevertheless, when adopting this approach, one must bear in mind that the conclusions of an ODE cannot be applied in a straightforward manner to the corresponding DNN, due to the fact that a numerical ODE may undergo several discretization approximations (e.g. forward/backward Euler approximations) that produce inconsistent results \cite{ascher1998computer}. 

Some researchers have sought to use existing knowledge of signal processing to design DNN-like networks 
to make the networks more comprehensive without sacrificing performance.  
This can often be achieved by replacing non-linear activation functions with interpretable non-linear operations in the form of non-linear transforms. Representative examples include the scattering transform \cite{mallat2012group,bruna2013invariant} and Saak transform \cite{kuo2018data}.
The scattering transform takes advantage of the wavelet transform and scattering operations in physics. The Saak transform employs statistical methods with invertible approximations. 

Network design was also inspired by optimization algorithms to solve ill-posed inverse problems \cite{gregor2010learning}. The un-rolling approach involves the systematic transformation of an iterative algorithm for an ill-posed inverse problem into a DNN. The number of iterations becomes the number of layers and the matrix in any given iteration is relaxed through the use of affine linear operations and activation functions. This makes it possible to infer the solution of the inverse problem using a DNN. This approach efficiently derives a network for an inverse problem, often achieving performance that is superior to the theoretical guarantees by the conventional inverse problem \cite{zhang2018ista,chan2019performance}; however, it does not provide  sufficient insight into the properties of DNNs  capable of solving the inverse problem. A through review of this topic can be found in \cite{monga2021algorithm}.


The un-rectifying method is closely tied to the problem-solving method used in piecewise functions, wherein the domain is partitioned into intervals to be analyzed separately. This approach takes advantage of the fact that a piecewise function is generally difficult to analyze as a whole, whereas it is usually a tractable function when the domain is restricted to a partitioned interval. 
When applying the un-rectifying method,  a point-wise CPWL activation function is replaced with a finite number of data-dependent linear mappings. This makes it possible to associate different inputs with different functions. 
The method replaces the point-wise CPWL activation function as data-dependent linear mapping as follows: 
\begin{align} \label{unrectify}
\rho (\bx)=  \bD^\rho_{\bx}(\bx),
\end{align}
where $\bD^\rho_{\bx}$ is the un-rectifying matrix for $\rho$ at $\bx$. If $\rho$ is the ReLU, then $\bD^\rho_{\bx}$ is a diagonal matrix with diagonal entries $\{0, 1\}$.The un-rectifying variables in the matrix provide crucial clues by which to characterize the function of the DNN. For example, comparing 
the following un-rectifying representation of $M_2 \circ \mathcal N_1^s$ with inputs $\bx$ and $\by$ respectively yields
\begin{align}
M_2 \circ \mathcal N_1^s(\bx) & = M_2 \bD_{M_1\bx} M_1 (\bx)  \label{N2sx}\\
M_2 \circ \mathcal N_1^s(\by) & = M_2 \bD_{M_1\by} M_1 (\by). \label{N2sy}
\end{align}
Note that the sole difference between (\ref{N2sx}) and (\ref{N2sy}) lies in the un-rectifying matrices $\bD_{M_1\bx}$ and $\bD_{M_1\by}$. 
Theoretical results for the series-connected networks, $ \mathcal N^s_L(\bx) 
= \varrho_L \circ M_L \circ\cdots\circ\varrho_1\circ M_1(\bx)
$ were derived using affine spline insights \cite{BaraniukPowerDiagramSubdiv} and the un-rectification approach\cite{Wen19,heinecke2020refinement}.

\section{DNNs and DAG representations} \label{DAGsec}

The class of DNNs addressed in this study is defined by specific activation functions, non-linear transformation, and underlying architecture.
Note that legitimate activation functions, non-linear transformations, and architectures should be analyzable and provide sufficient generalizability to cover all DNNs in common use. 

Activation functions and non-linear transformations are both considered functions; however, we differentiate between them because of the different ways that they are treated under un-rectifying analysis. 
A non-linear transformation is a function in the conventional sense when mapping $\R^n$ to $\R^m$, in which different inputs are evaluated using the same function. This differs from activation functions in which different inputs can be associated with different functions. For example, for ReLU $\R \rightarrow \R$, un-rectifying considers $\text{ReLU} x =dx $ where $d \in \{0, 1\}$ as two functions, depending on whether $x > 0$ where $d=1$ or $x \leq 0$ where $d=0$.

\subsection{Activation functions and non-linear transformations}

In this paper, we focus on activation functions that can be expressed as networks of point-wise CPWL activation functions $\rho$. Based on this assumption and the following lemma, we assert that the activation functions of  concern are ReLU networks.

\begin{citedlem}\cite{arora2016understanding} \label{RELU1}
Any point-wise CPWL activation function $\rho: \R \rightarrow \R$ of $m$ pieces can be expressed as follows:
\begin{align} \label{PWm}
\rho(x) = \sum_{i=1}^m r_i \text{ReLU}(x - a_i) + l_i \text{ReLU}(t_i - x) = \sum_{i \in I^+} r_i \text{ReLU}(x - a_i) +  \sum_{i \in I^-} l_i \text{ReLU}(t_i - x)
\end{align}
where $l_i$ and $r_i$ indicate the slopes of segments and $a_i$ and $t_i$ are breakpoints of the corresponding segments. 
\end{citedlem}
Note that the max-pooling operation (arguably the most popular non-linear pooling operation), which outputs the largest value in the block and maintains the selected location \cite{goodfellow2013maxout} is also a ReLU network. The max-pooling of a block of any size can be recursively derived using the max-pooling of a block of size $2$. For example, let $\max_4$ and $\max_2$ respectively denote the max-pooling of blocks of sizes $4$ and $2$. Then, ${\max}_4(x_1, x_2, x_3, x_4) = {\max}_2 ({\max}_2(x_1, x_2), {\max}_2(x_3, x_4))$ and ${\max}_5(x_1, x_2, x_3, x_4, x_5) = {\max}_2 ({\max}_4(x_1, x_2, x_3, x_4), x_5)$. The max-pooling of a block of size $2$ can be expressed as follows:
\begin{align} 
{\max}_2(\bx = [x_1 \;  x_2]^\top) & = \frac{x_1 + x_2}{2} + \frac{|x_1 - x_2|}{2}  \nonumber \\
& =\frac{1}{2} [1 \; 1 \; 1] \tilde \rho \begin{bmatrix} 1 & 1 \\ 1 & -1 \\ -1 & 1 \end{bmatrix} \bx, \label{maxpooling2}
\end{align}
where $\tilde \rho \in \R^{3 \times 3}$ is 
\begin{align*}
\tilde \rho = \begin{bmatrix} 1 & 0 & 0\\ 0 & \text{ReLU} & 0 \\ 0 & 0  &\text{ReLU}  \end{bmatrix}.
\end{align*}

We make two assumptions pertaining to activation function $\rho$ considered in this paper: \\
(A1) $\rho:\R\rightarrow \R$ can be expressed as (\ref{PWm}).

This assumption guarantees that for any input $\bx$,  the layer of $\rho: \R^l \rightarrow \R^l$ can be associated with diagonal un-rectifying matrix $\bD_{\bx}^\rho$ with real-valued diagonal entries, where the value of the $l$-th diagonal entry is 
$ \sum_{i \in I_l^+} r_{l,i} + \sum_{j \in I_l^{-}} l_{l,j}$, 
where $I_l^+$ and $I_l^{-}$ denote the sets in which ReLUs are active.

(A2) There exists a bound $d_{\rho} > 0$ for any activation function $\rho$ for any input $\bx$. This corresponds to the assumption that 
\begin{align} \label{uniformboundact}
\|\bD^{\rho}_{\bx}\|_2 = \max_l |\sum_{i \in I_l^+} r_{l,i} + \sum_{j \in I_l^{-}} l_{l,j}| \leq d_{\rho}.
\end{align}
This assumption shall be used to establish the stability of a network against input perturbations.


The outputs of a non-linear transformation layer can be interpreted as coefficient vectors related to that domain of the transformation. We make the following assumption pertaining to non-linear transformation $\sigma$ addressed in the current paper. \\
(A3) There exists a uniform Lipschitz constant bound $d_{\sigma} > 0$ with respect to $\ell_2$-norm for any non-linear transformation function $\sigma$ for any inputs $\bx$ and $\by$ in $\R^n$:
\begin{align} \label{uniformboundfor}
\| \sigma(\bx) - \sigma(\by) \|_2  \leq d_{\sigma} \| \bx - \by \|_2.
\end{align}
This assumption shall be used to establish the stability of a network against input perturbations. Sigmoid and tanh functions are 1-Lipschitz \cite{krizhevsky2012imagenet}. The softmax layer from $\R^n$ to $\R^n$ is defined as $x_i \rightarrow \frac{\exp^{\lambda x_i}}{\sum_{j=1}^n \exp^{\lambda x_j}}$ where $\lambda$ is the inverse temperature constant. The output is the estimated probability distribution of the input vector in the simplex of $\R^n$ and $i=1, \cdots, n$.
Softmax function $\text{Softmax}_{\lambda}$ persists as $\lambda$-Lipschitz \cite{gao2017properties}, as follows:
\begin{align*}
\| \text{Softmax}_{\lambda}(\bx) - \text{Softmax}_{\lambda}(\by) \|_2 \leq \lambda \| \bx - \by \|_2.
\end{align*}


\subsection{Proposed axiomatic method}

Let $\cK$ denote the class of DNNs that can be constructed using the following axiomatic method with activation functions that satisfy (A1) and (A2) and non-linear transformations that satisfy (A3).
This axiomatic method employs three atomic operations (O1-O3),  the basic set $\cB \subseteq \cK$, and a regulatory rule (R) describing the legitimate method by which to apply an atomic operation to elements in $\cK$ in order to yield another element in $\cK$. 

Basis set $\cB$ comprises the following operations:
\begin{align} 
\cB = \{ \bI, \bL, M, \Gamma_{\rho}, \rho M, \Gamma_{\sigma}, \sigma M \},
\end{align}
where $\bI$ denotes the identify operation; $\bL$ denotes any finite dimensional linear mapping with a bounded spectral-norms; $M = (\bL, \bb)$ denotes any affine linear mapping where $\bL$ and $\bb$ respectively refer to the linear and bias terms; $\Gamma_{\rho}$ denotes activation functions satisfying (A1) and (A2); $\rho M$ denotes functions with $\rho \in \Gamma_{\rho}$; $\Gamma_{\sigma}$ denotes non-linear transformations satisfying (A3); and $\sigma M$ denotes functions with $\sigma \in \Gamma_{\sigma}$.


Assumptions pertaining to $\Gamma_{\rho}$ and $\Gamma_{\sigma}$ are combined to obtain the following: \\
(A) The assumption of uniform bounding is based on the existence of a uniform bound, where $\infty > d > 0$ for any activation function $\rho \in \Gamma_{\rho}$ for any input $\bx$ and any non-linear transformation $\sigma \in \Gamma_{\sigma}$, such that  
\begin{align} \label{uniformbound}
d \ge \max\{ d_{\rho}, d_{\sigma}\}.
\end{align}

\begin{figure}[th]
\centering
\subfigure[A series-connection]{\includegraphics[width=\twohalf\textwidth]{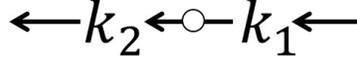}}\\
\subfigure[Concatenation]{\includegraphics[width=\twohalf\textwidth]{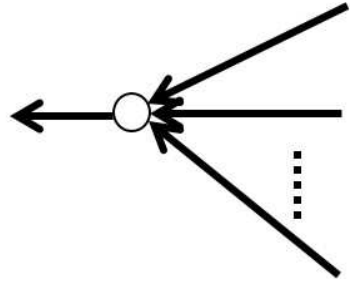}}\\
\subfigure[Duplication]{\includegraphics[width=\twohalf\textwidth]{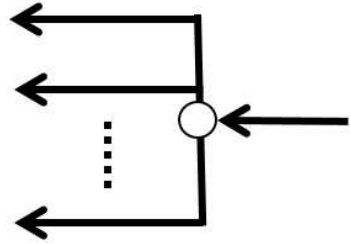}} 
\caption{Graphical representation of atomic operations O1-O3, where functions attached to arcs of concatenation and duplication are identify operation $\bI$ (omitted for brevity).} 
\label{figbasic}
\end{figure}

Let $\chi$ denote the input space for any elements in $\cB$. The results of the following atomic operations belong to $\cK$. The corresponding DAG representations are depicted in Figure \ref{figbasic} (a reshaping of input or output vectors is implicitly applied at nodes to validate these operations).
\begin{enumerate}
\item[O1.] Series-connection ($\circ$): We combine $\cB$ and $\cK$ by letting the output of $k_1 \in \cK$ be the input of $k_2  \in \cB$, where
\begin{align*}\cB \circ \cK : k_2 \circ k_1:\bx \rightarrow k_2(k_1 \bx). \end{align*}
\item[O2.] Concatenation: We combine multi-channel inputs $\bx_i \in \chi$ into a vector,  as follows:
\begin{align*}
\text{concatenation}: \{\bx_1, \cdots, \bx_m\} \rightarrow [\bx_1^\top \; \cdots \bx_m^\top]^\top.
\end{align*}
\item[O3.] Duplication: We duplicate an input to generate $m$ copies of itself, as follows: 
\begin{align*}
\text{duplication}: \bx \in \chi \rightarrow \begin{bmatrix}
\bI \\
\vdots \\
\bI 
\end{bmatrix} \bx.
\end{align*}
\end{enumerate}

From the basic set and O1-O3, regulatory rule R generates other elements in $\cK$ by regulating the application of atomic operations on $\cK$. The aim of R is to obtain DNNs representable as DAGs; therefore,  this rule precludes the generation of graphs that contain loops.

\begin{itemize}
\item[R.] DAG-closure: We apply O1-O3 to $\cK$ in accordance with 
\begin{align*}
R: \cK \rightarrow \cK.
\end{align*}
\end{itemize}
The DAGs of $\cK$ comprise nodes and arcs, each of which belongs to one of operations O1-O3. Rule R mandates that any member in $\cK$ can be represented as a DAG, in which arcs are associated with members in $\cB$ and nodes coordinate the inlets and outlets of arcs. The rule pertaining to the retention of DAGs after operations on DAGs is crucial to our analysis, based on the fact that nodes in a DAG can then be ordered (see Section \ref{commonprop}). To achieve a more comprehensive understanding, we use figures to express DAGs. Nevertheless, a formal definition of a DAG must comprise triplets of nodes, arcs, and functions associated with arcs. An arc can be described as $(v_i, k, v_o)$ where $v_i$ and $v_o$ respectively refer to the input and output nodes of the arc, and $k \in \cB$ is the function associated with the arc.

We provide the following definition for the class of DNNs considered in this paper.
\begin{citeddef}
DNNs constructed using the axiomatic method involving point-wise CPWL activation functions and non-linear transformations, which together satisfy assumption (A), are referred to as general DNNs (denoted as $\cK$).
\end{citeddef}
Note that DNNs comprise hidden layers and an output layer. For the remainder of this paper, we do not consider the output layers in DNNs, because adding a layer of continuous output functions does not alter our conclusion.


\subsection{Useful modules}

Generally, the construction of DNN networks is based on modules. 
Below, we illustrate some useful modules in pragmatic applications of DNNs is in $\cK$.

\begin{figure}[th]
\begin{center}
\subfigure[]{\includegraphics[width=0.25\textwidth]{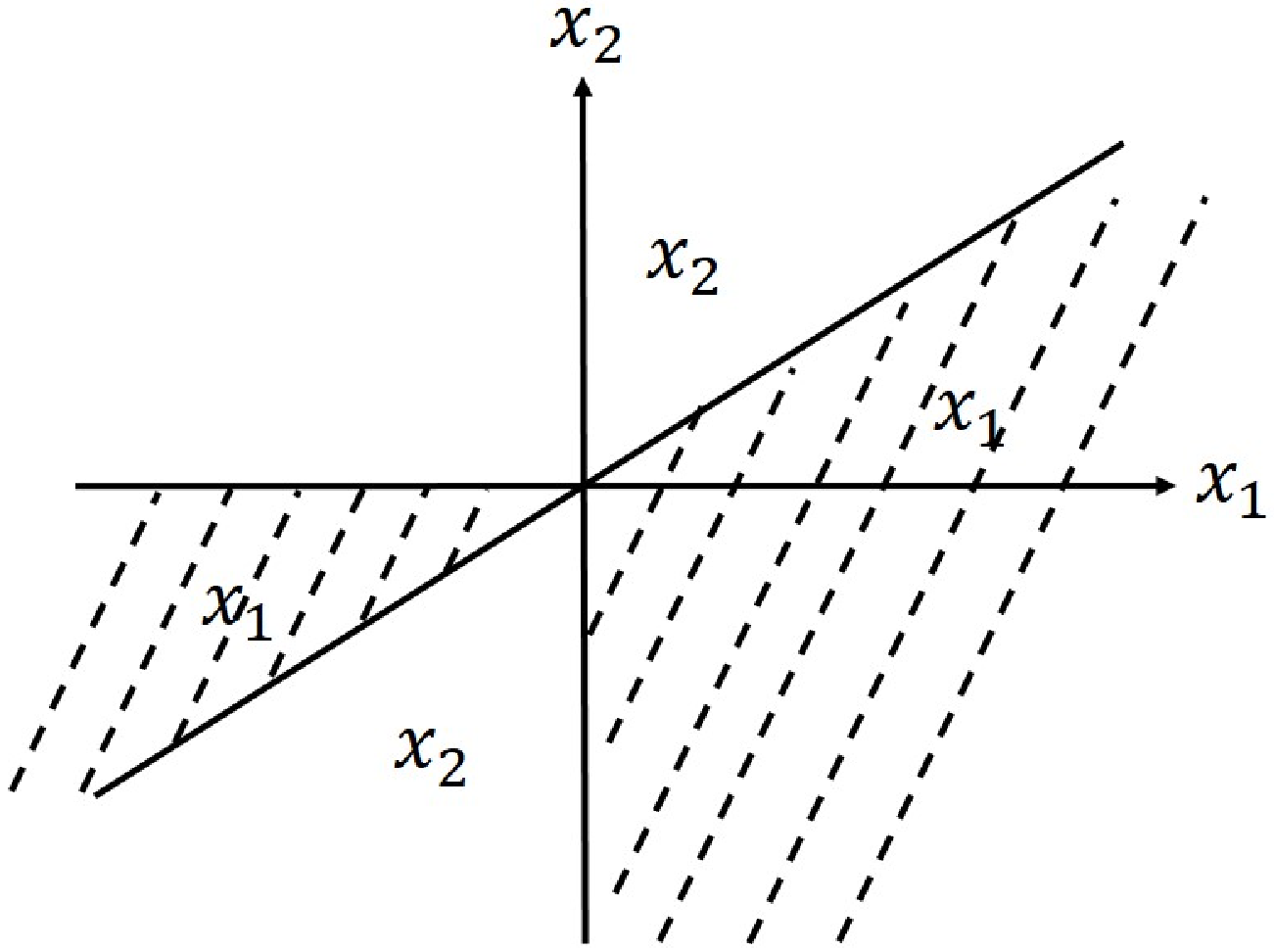}}
\subfigure[]{\includegraphics[width=0.25\columnwidth]{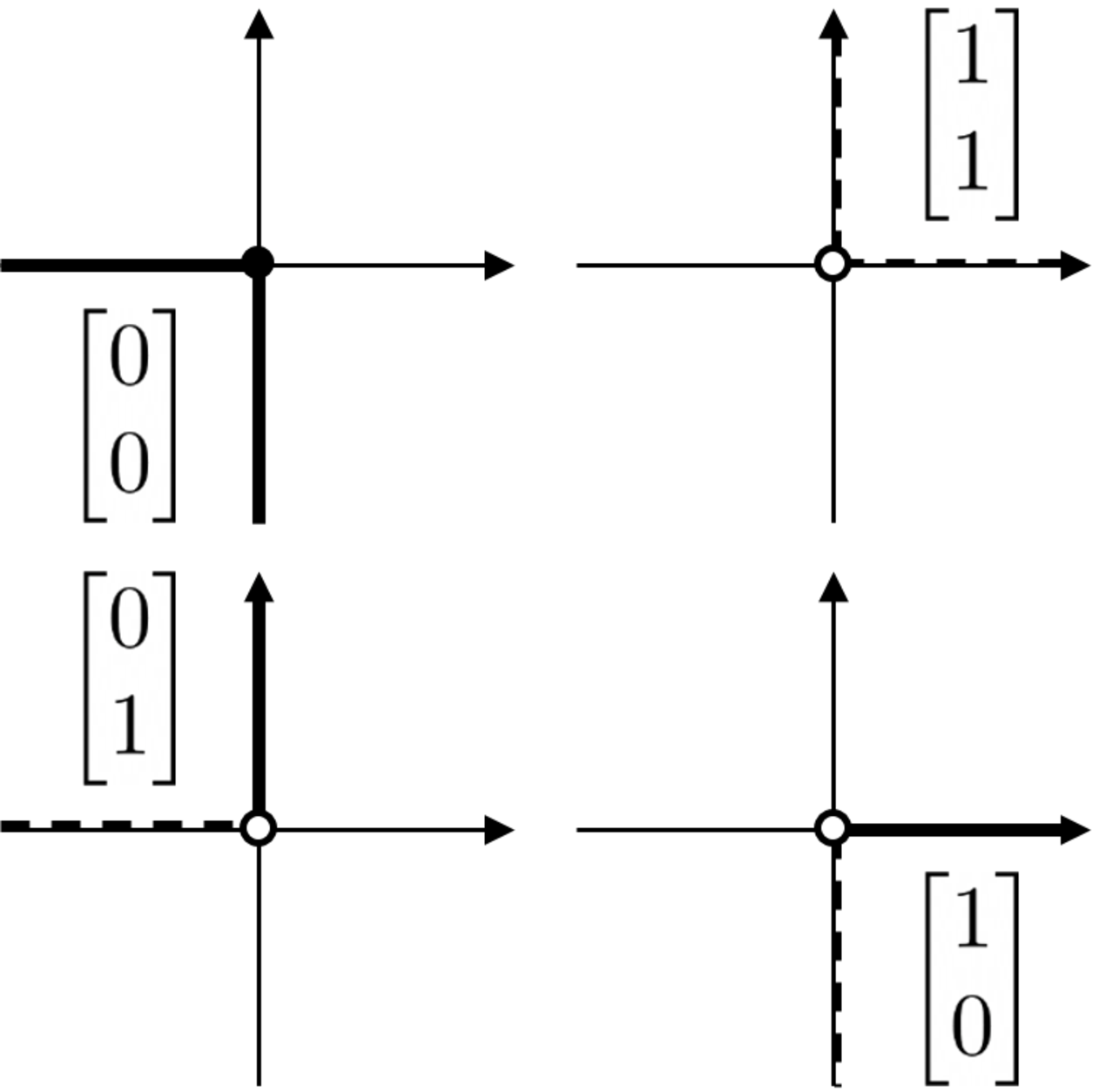}}
\subfigure[]{\includegraphics[width=0.25\columnwidth]{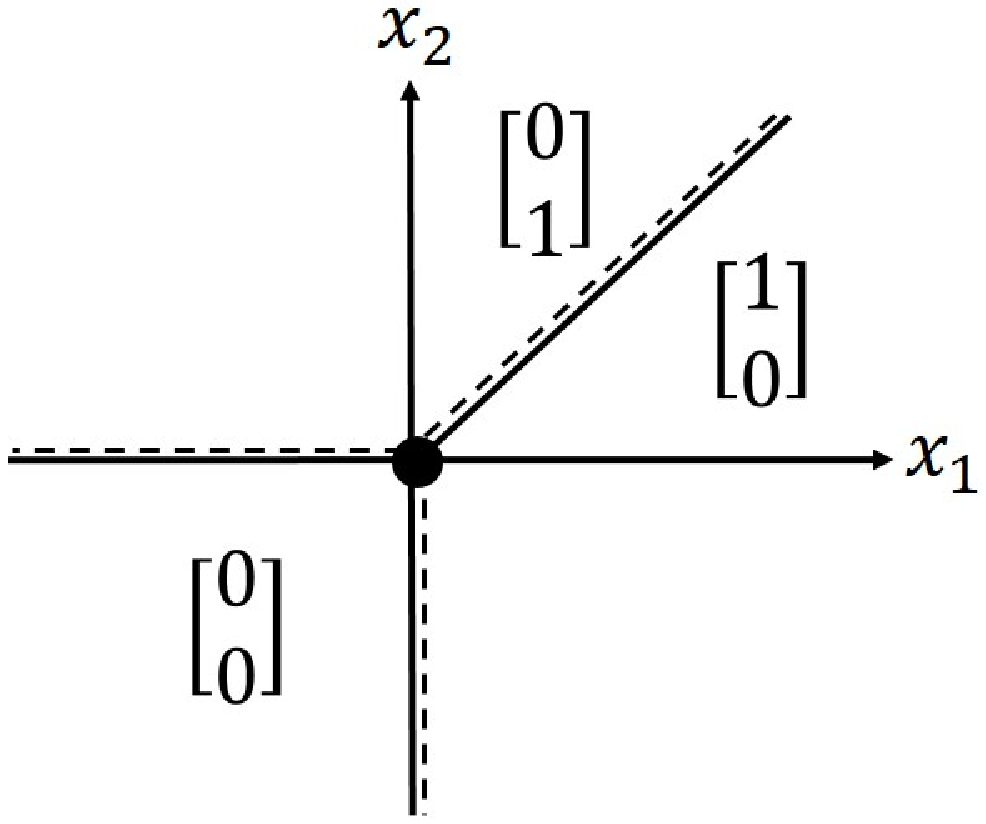}}
\end{center}
\caption{
Comparisons of Max-pooling, ReLU, and MaxLU$_2$ partitioning of $\R^2$ where vectors in (b) and (c)  are diagonal vectors of the un-rectifying matrices of ReLU and MAXLU layers and dashed lines denote open region boundaries: (a) $\max_2$ partitions $\R^2$ into two regions; (b) ReLU partitions $\R^2$ into four polygons; and  (c) MaxLU$_2$ partitions $\R^2$ into three polygons. Note that the region boundary in the third quadrant of (a) is removed in (c). } \label{PW}
\end{figure}

1) MaxLU module: Pooling is an operation that reduces the dimensionality of an input block. The operation can be linear or non-linear. For example, average-pooling is linear (outputs the average of the block), whereas max-pooling is non-linear. Max-pooling is usually implemented in conjunction with the ReLU layer (i.e., maxpooling $\circ$ ReLU) \cite{goodfellow2016deep} to obtain MaxLU module. The following analysis is based on the MaxLU function of block-size $2$ using un-rectifying (the MaxLU function of another block-size can be recurrently derived using the MaxLU of block-size $2$ and analysed in a similar manner) \cite{Wen19}. 
The MaxLU$_2:\R^2 \rightarrow \R$ is a CPWL activation function that partitions $\R^2$ into three polygons. Representing the MaxLU layer with un-rectifying, we obtain the following:
\begin{align*}
\text{MaxLU}_2(\bx= [x_1 \; x_2]^\top) = [1 \;  1] \bD^\sigma_{\bx} \bx.
\end{align*}
$\bD^\sigma_{\bx} \in \mathbb R^{2 \times 2}$ is a diagonal matrix with entries $\{0, 1\}$ and 
\begin{align*}
\text{diag}(\bD^\sigma_{\bx})= 
\begin{cases}
[1 \; 0]^\top  \text{ when $x_1 \geq x_2$ and $x_1 > 0$} \\
[0 \; 1]^\top \text{ when $x_2 > 0$ and $x_2 > x_1$} \\
[0 \;  0]^\top \text{ otherwise.}
\end{cases}
\end{align*}
Figure \ref{PW} compares the domain partition of Max-pooling, ReLU, and MaxLU$_2$.

%

2) Series module: This module is a composition of $\cK$ and $\cK$ (denoted as $\cK \circ \cK$) (compared to operation O1, $\cB \circ \cK$). 

\begin{figure}
\begin{center}
\includegraphics[width=0.32\textwidth]{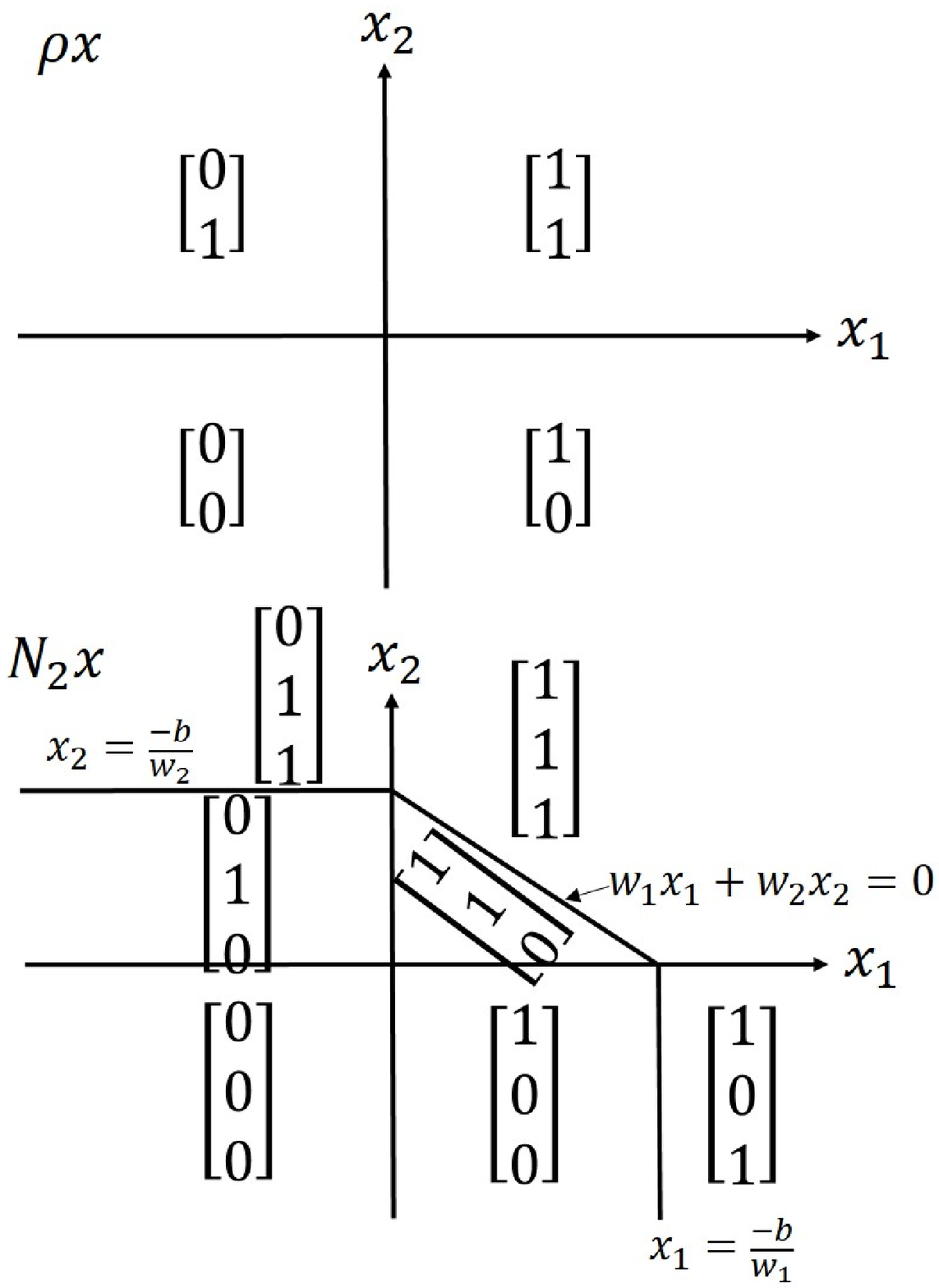} \hspace{0.1in} \includegraphics[width=0.35\textwidth]{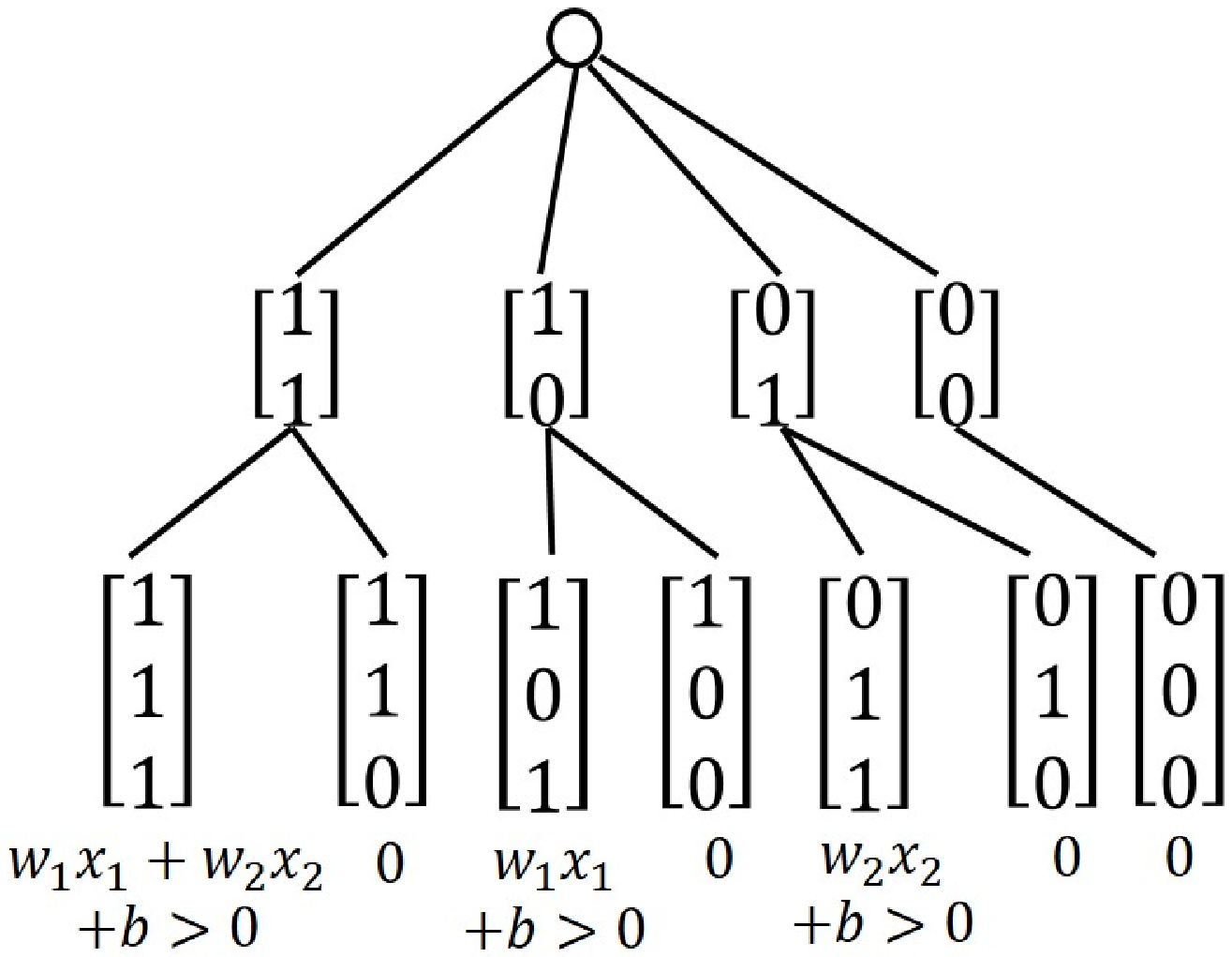}
\end{center}
\caption{(Top-left) $\R^2$ is partitioned using ReLU$_1$ and (bottom-left) refined using ReLU$_2M_1$. (Right) Tree partition on $\R^2$ from composition of ReLUs, where the function with domain on a region is indicated beneath the leaf and regions are named according to vectors obtained by stacking diagonal elements of the un-rectifying matrix of ReLU$_1$ over the un-rectifying matrix of ReLU$_2$.} \label{composition}
\end{figure}

The series-connected network $ \mathcal N^s_L(\bx) 
= \varrho_L \circ M_L \circ\cdots\circ\varrho_1\circ M_1(\bx)
$ is derived using a sequence of series modules.
We consider that $\rho_i$ are ReLU. Theoretical results for this module with MaxLU activation functions can refer to \cite{Wen19}. We first present an illustrative example of $\mathcal N^s_2$ and then extend our analysis to $\mathcal N^s_L$. 
Note that input space $\chi$ is partitioned into a finite number of regions using $(\text{ReLU}_1) M_1$, where $M_1$ is an affine mapping. The partition is denoted as $\mathbb P_1$.
The composition of $(\text{ReLU}_2) M_2$ and $(\text{ReLU}_1) M_1$ (i.e., $\text{ReLU}_2 M_2 \circ \text{ReLU}_1 M_1$) refines $\mathbb P_1$, such that the resulting partition can be denoted as $\mathbb P_2$. Figure \ref{composition} presents a tree partition of $\R^2$ using $(\text{ReLU}_2)M_2 (\text{ReLU}_1): \mathbb R^2 \rightarrow \mathbb R^1$, where
$M_2 \bx = w_1 x_1 + w_2 x_2 + b$, in which $\bx = [x_1\; x_2]^\top$, $w_1, w_2  \ge 0$, and $b \leq 0$. 
The affine linear functions over $\mathbb P_2$ can be expressed as $\bD^2 M_2 \bD^1 \R^2$, where $\bD^1$ and $\bD^2$ respectively indicates un-rectifying matrices of ReLU$_1$ and ReLU$_2$ using $\bx$ and $M_2 \text{ReLU}_1 \bx$ as inputs.

For a series-connected network $\mathcal N^s_{i}$, we let $(\mathbb P_{i}, \mathbb A_{i})$ denote the partition and corresponding functions. 
The relationship between $(\mathbb P_{L}, \mathbb A_{L})$ and  $(\mathbb P_{L-1}, \mathbb A_{L-1})$  is presented as follows.
\begin{citedlem}[\cite{Wen19}]\label{seriesprop}
Let $(\mathbb P_{i} = \{P_{i,k}\}, \mathbb A_{i} =\{\bA_{i,k}\})$ denote the partition of the input space $\chi$ and the collection of affine linear functions of $\mathcal N_i^s$. Further, let the domain of the affine linear function $\bA_{i,k}$ be $P_{i,k}$. 
Then, \\
(i) $\mathbb P_L$ refines $\mathbb P_{L-1}$ \footnote{Any partition region in $\mathcal P_L$ can be  subsumed to one and only one partition region in $\mathbb P_{L-1}$}. \\
(ii) The affine linear mappings of $\mathcal N_L^s= \text{ReLU}_L M_L \mathcal N^s_{L-1}$ can be expressed as
$\mathbb A_L \chi = \bD^L M_L  \mathbb A_{L-1} \chi$, where $\bD^L$ is an un-rectifying matrix of  ReLU$_L$. That is, if $\bA_{L, i} \in \mathbb A_L$, then there must be a $j$ in which $P_{L,i} \subseteq P_{L-1,j}$, such that $\bA_{L,i} P_{L,i} = \bD^{L,i} M_L \bA_{L-1, j} P_{L, i}$ where the un-rectifying matrix $\bD^{L,i}$ depends on $M_L \bA_{L-1, j} P_{L, i}$.
\end{citedlem}

\begin{figure}[tp]
\centering
\subfigure[Parallel module]{\includegraphics[width=\twohalf\textwidth]{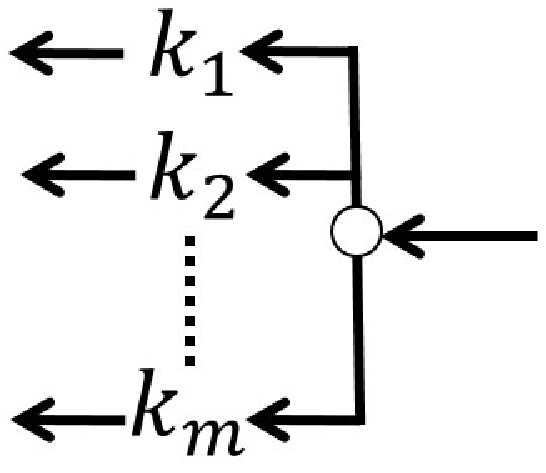}}
\subfigure[Fusion module]{\includegraphics[width=\twohalf\textwidth]{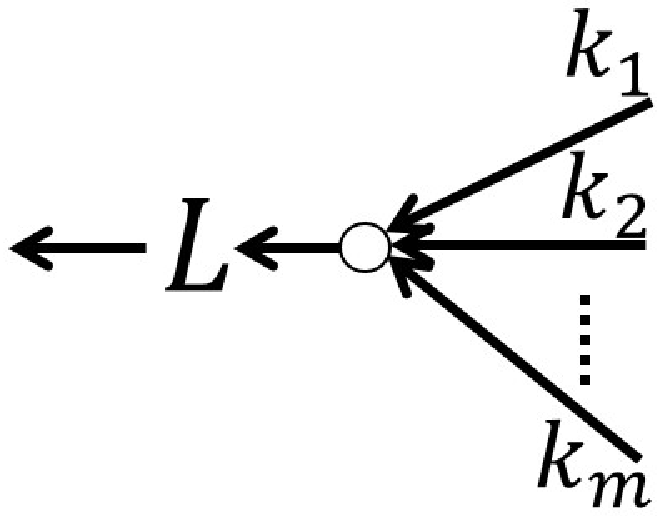}}
\caption{Graphical representations of (a) parallel module (which becomes a duplication operation when $k_i = \bI$), and (b) fusion module (which becomes a concatenation operation when $\bL = \bI$ and $k_i = \bI$). }\label{figmodulus}
\end{figure}

3) Parallel module (Figure \ref{figmodulus}(a)): This module is a composition comprising an element in $\cK$ to each output of the duplication operation, denoted as follows:
\begin{align*}
\text{(parallel)}  \doteq  \cK \circ \text{(duplication)}.
\end{align*}
When expressed in matrix form, we obtain the following: 
\begin{align*}
\bx \rightarrow 
\begin{bmatrix}
\bI \\
\vdots \\
\bI 
\end{bmatrix} \bx \rightarrow
\begin{bmatrix}
k_1 \\
\vdots \\
k_m 
\end{bmatrix}\bx.
\end{align*} 
In the literature on DNNs, this module is also referred to as a multi-filter ($k_i$ is typically a filtering operation followed by an activation function) or multi-channel.

4) Fusion module (Figure \ref{figmodulus}(b)): This type of module can be used to combine different parts of a DNN and uses linear operations to fuse inputs. The module is denoted as follows:
\begin{align}
\text{(fusion)} \doteq 
\bL \circ \text{(concatenation)}.
\end{align}
In matrix form, we obtain the following:
\begin{align*}
\bx \rightarrow \{k_1 \bx_1, \cdots, k_m \bx_m\} \rightarrow  \bL\begin{bmatrix} k_1\bx_1\\ \vdots \\ k_m \bx_m\end{bmatrix}.
\end{align*}
Note that a non-linear fusion can be obtained by applying a composition of $\rho M$/$\rho$ to the fusion module in which $\bL = \bI$. 

We denote the domain partition of $\chi$ associated with the $i$-th channel as $\mathbb P_i = \{ R_{ij_i}| j_i=1, \cdots n_i\}$, where $n_i$ is the number of partition regions. The partition $\mathbb P$ of $\chi$ induced by the fusion module can be expressed as the union of non-empty intersection of partition regions in $\mathbb P_i$ for all $i$, as follows:
\begin{align} \label{fusiondom}
\mathbb P & \doteq \mathbb P_1 \cap \mathbb P_2 \cap \cdots \cap \mathbb P_m \\
& = \cup_{j_1, \cdots, j_m} ( R_{1j_1} \cap  R_{2j_2}  \cdots \cap R_{mj_m} \neq \emptyset). \nonumber
\end{align}
Any partition region in $\mathbb P$ is contained in precisely one region in any $\mathbb P_i$. In other words, $\mathbb P$ is a refinement of $\mathbb P_i$ for $i=1, \cdots, m$.
An obvious bound for partition regions of $\mathbb P$ is $\prod_{i=1}^m n_i$.

We let $\{A_{ij_i}| j_i= 1\cdots n_i\}$ denote the affine mappings associated with the $i$-th channel, and let $A_{i j_i}$ be the affine mapping with domain restricted to partition region $R_{ij_i}$.
The affine mapping of the fusion module over partition region $R_{1j_1} \cap R_{2j_2} \cap \cdots \cap R_{mj_m} \neq \emptyset$ is derived as follows: 
\begin{align} \label{linearfusionmap}
R_{1j_1} \cap R_{2j_2} \cap \cdots \cap R_{mj_m} \neq \emptyset \rightarrow 
\bL \begin{bmatrix}
A_{1j_1} \\
\vdots \\
A_{mj_m}
\end{bmatrix}.
\end{align}
For the sake of convenience, (\ref{fusiondom}) and (\ref{linearfusionmap}) are summarized in the following lemma.
\begin{citedlem} \label{fusionlem}
Suppose that a fusion module comprises $m$ channels. 
Let $\mathbb P_{i}$ denote the partition associated with the $i$-th channel of the module, and let $\mathbb P$ denote the partition of the fusion module. Then, $\mathbb P$ is a refinement of $\mathbb P_i$ for $i= 1, \cdots, m$. Moreover, let $\mathbb A$ denote the collection of affine linear mappings over $\mathbb P$. Thus, the affine linear mapping over a partition region of $\mathbb P$ can be obtained in accordance with (\ref{linearfusionmap}).
\end{citedlem}

\begin{exam}
Figure \ref{figparallel} illustrates the fusion of two channels, as follows:
\begin{align*}
\mathcal M: \R^2 \rightarrow \bL \begin{bmatrix} \rho_1 M_1 \\ \rho_2 M_2 \end{bmatrix},
\end{align*}
where $M_1, M_2: \R^2 \rightarrow \R^2$ and $\rho_1$ and $\rho_2$ are ReLUs. The partition induced by $\mathcal M$ comprises eight polytopes, each of which is associated with an affine linear mapping. 
\begin{figure}[tp]
\centering
\subfigure[$\mathbb P_1$: partition by $\rho_1 M_1.$]{
\includegraphics[width=\twohalf\textwidth]{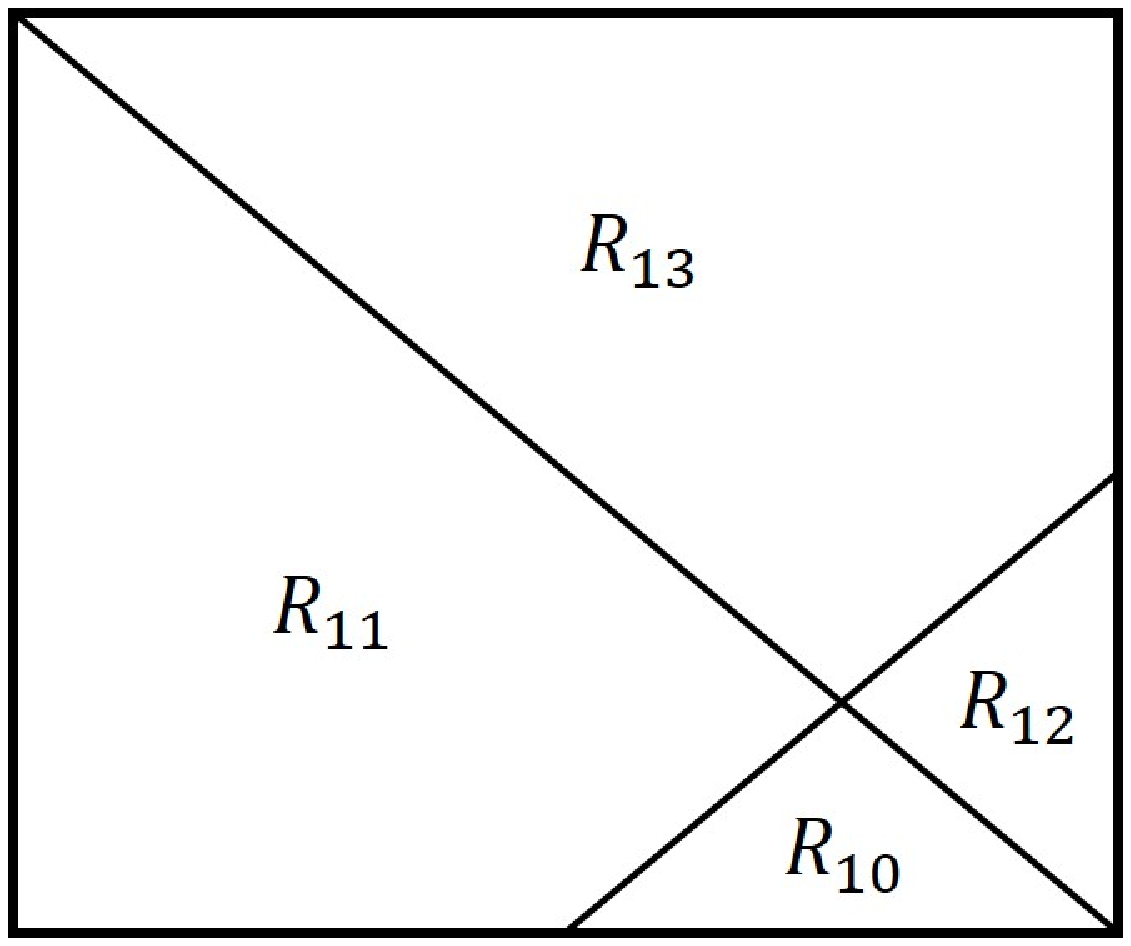}}\hspace{0.1in}
\subfigure[$\mathbb P_2$: partition by $\rho_2 M_2$.] {
\includegraphics[width=\twohalf\textwidth]{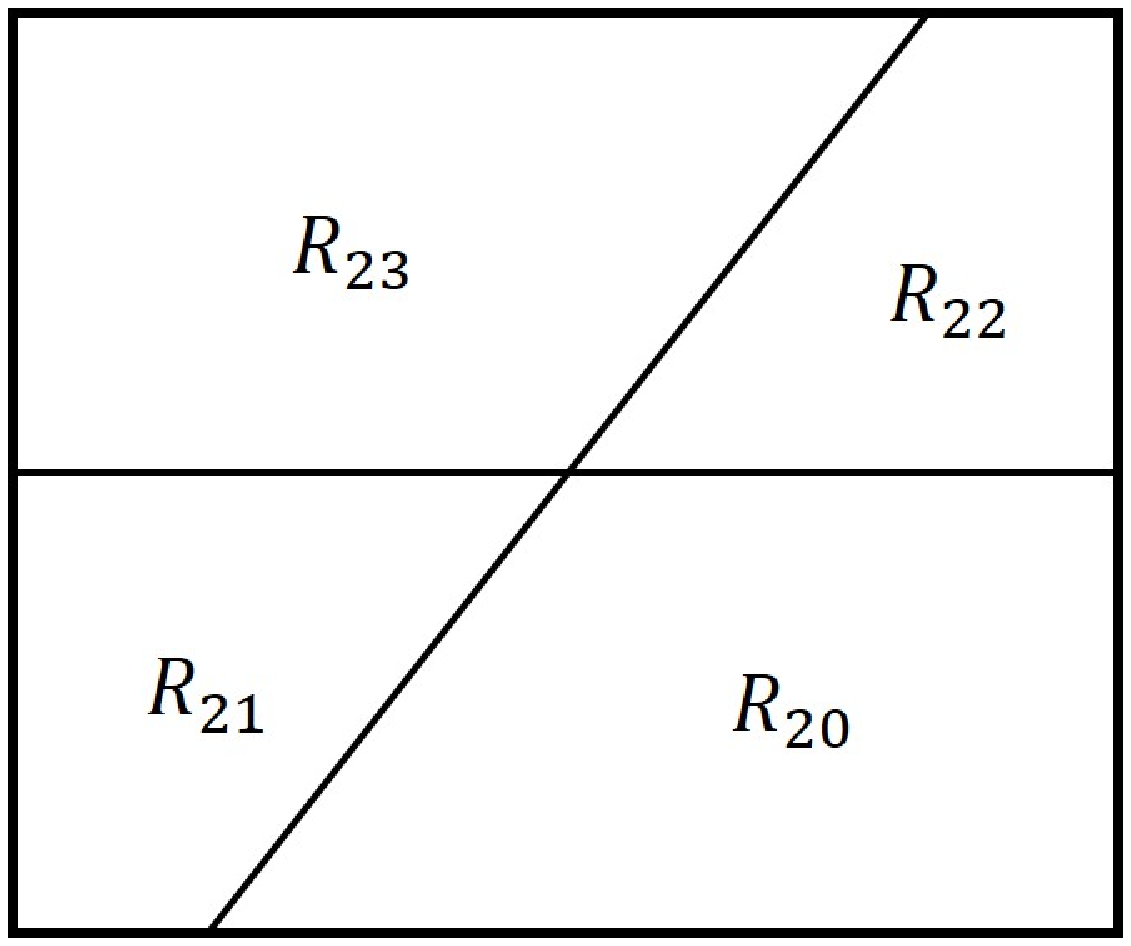}}\\
\vspace{0.1in}
\subfigure[$\mathbb P$: partition by fusion module.] {
\includegraphics[width=\twohalf\textwidth]{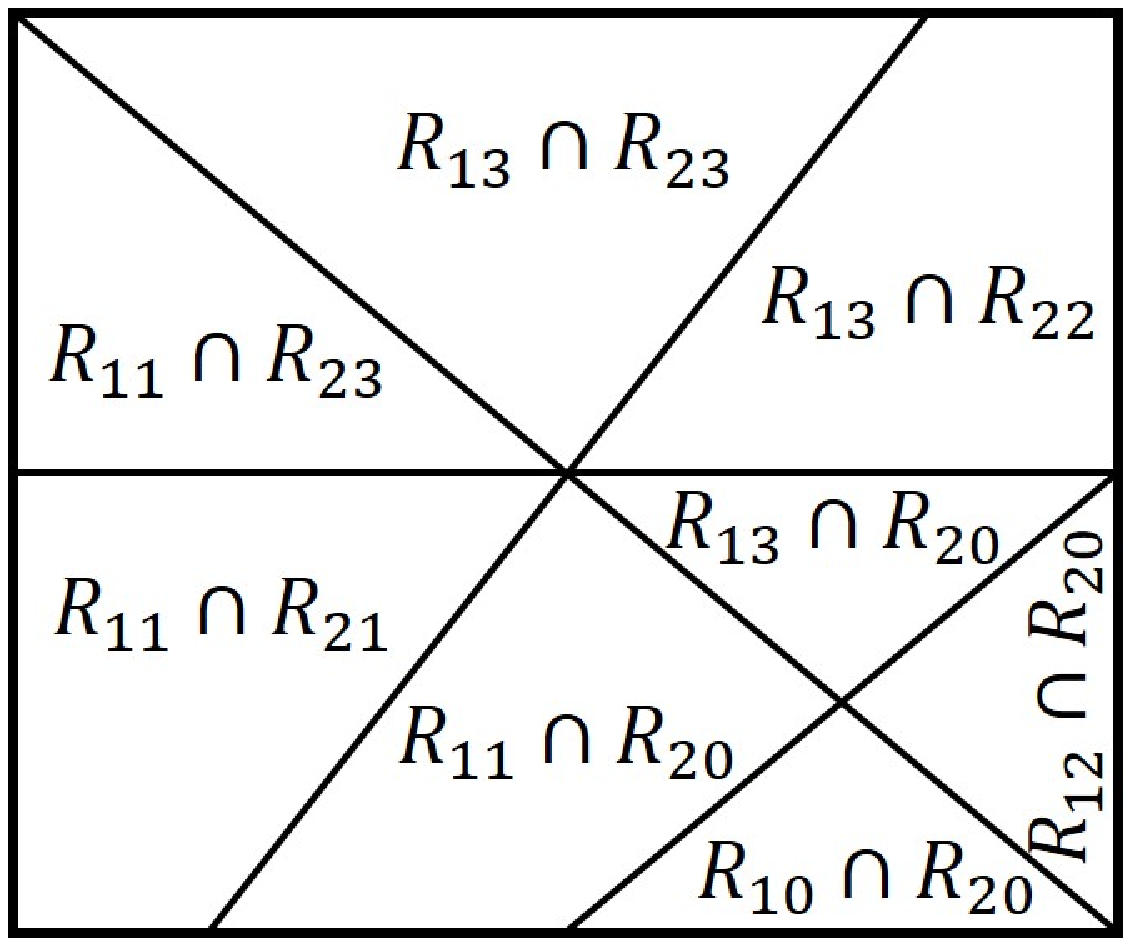}}
\caption{Fusion of two channels in which each channel partitions $\R^2$ into four regions: (a) $\mathbb P_1$ is the partition due to $\rho_1 M_1$, where $M_1: \R^2 \rightarrow \R^2$; (b) $\mathbb P_2$ is the partition due to $\rho_2 M_2$, where $M_2: \R^2 \rightarrow \R^2$; and (c) $\mathbb P$ is a refinement of $\mathbb P_1$ and $\mathbb P_2$. }
\label{figparallel}
\end{figure}
\end{exam}

\begin{figure}[!h]
\begin{center}
\subfigure[]{\includegraphics[width=0.6\textwidth]{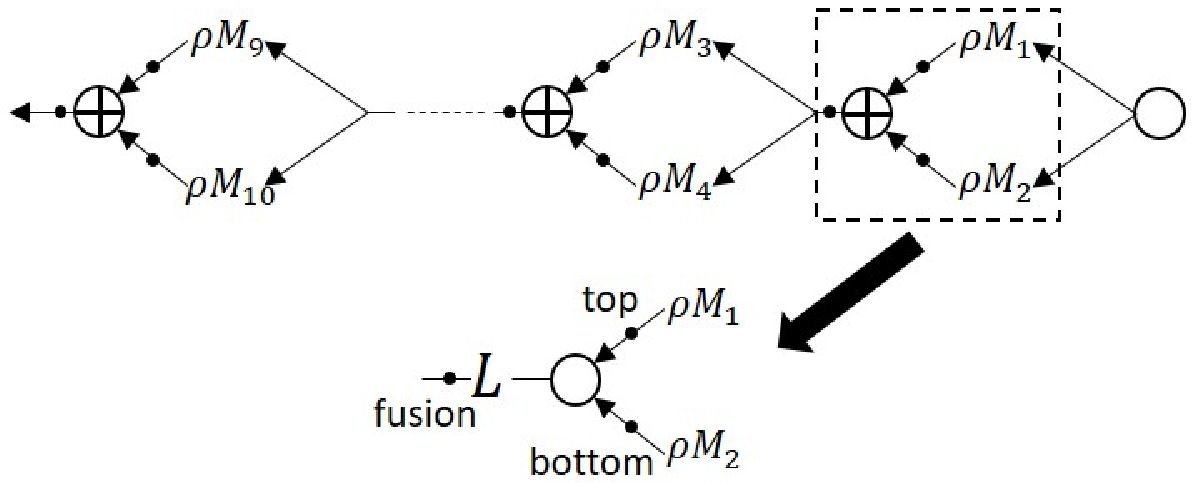}} \\
\subfigure[]{\includegraphics[width=0.43\textwidth]{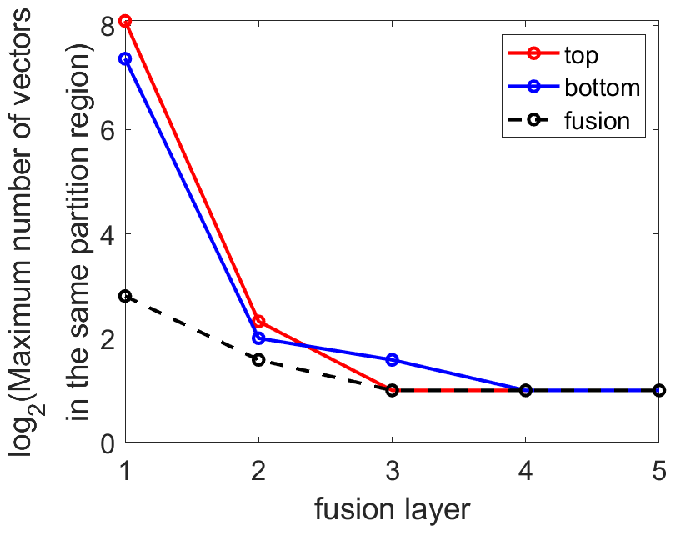}}
\subfigure[]{\includegraphics[width=0.43\textwidth]{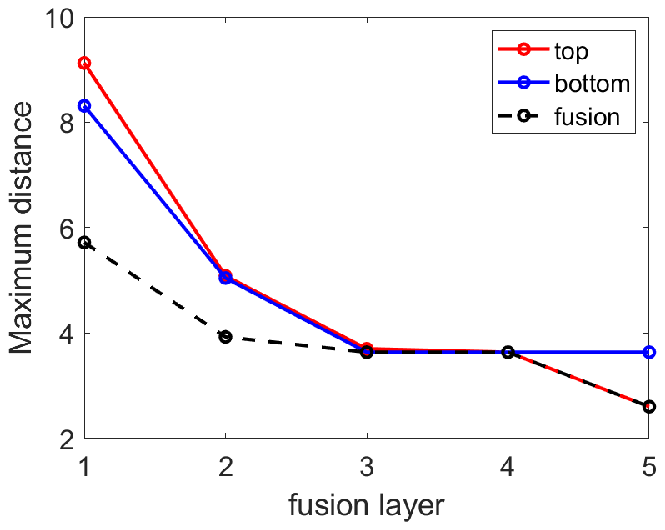}}
\subfigure[]{\includegraphics[width=0.43\textwidth]{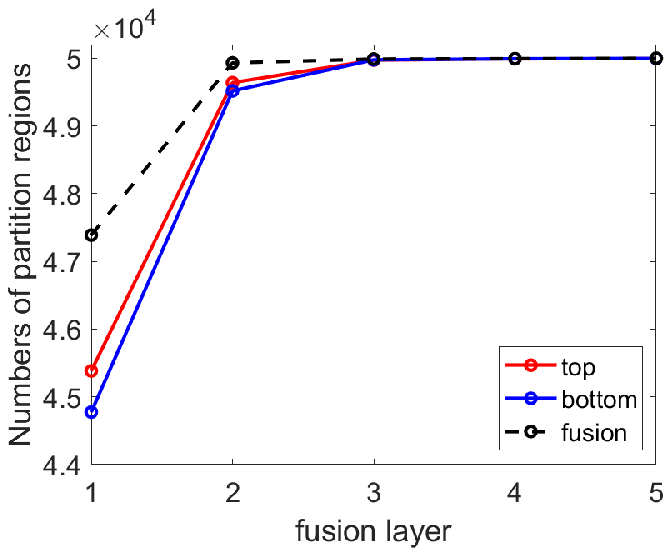}}
\end{center}\caption{Simulation of partition refinement using fusion modules with $50,000$ random points$(14 \times 1)$ as inputs with entries sampled independently and identically distributed (i.i.d.) from standard normal distribution. (a) Network comprising five layers of fusion modules, each of which comprises three channels (top, bottom, fusion). The dimensions of weight matrix and bias of $M_i$ in the top and bottom channels are $14 \times 14$ and $14 \times 1$, respectively, with coefficients in $M_i$ sampled i.i.d. from standard normal distribution and $\bL=\begin{bmatrix} \bI & \bI \end{bmatrix}$; (b) The number of points in partition regions containing at least two elements versus fusion layers; (c) The maximum distance between pairs of points located in the same partition region versus fusion layers; (d) The number of partition regions versus fusion layers. The fact that in (b) and (c) the curves corresponding to fusion channels are beneath those of the other channels and that in (d) the curve of fusion channel is above the other channels are consistent with the analysis that the partition at fusion channel is finer than those at the bottom and top channels.}
\label{fig:ex_partition}
\end{figure}

Figure \ref{fig:ex_partition} illustrates the refinement of partitions in the network in Figure  \ref{fig:ex_partition}(a), which was obtained through a series-connection of five fusion layers. Each fusion layer involves a fusion module derived by concatenation of inputs $\text{ReLU}M_1$ (top) and $\text{ReLU}M_2$ (bottom). The result of the concatenation is subsequently input to linear function $\bL=\begin{bmatrix} \bI & \bI \end{bmatrix}$ (fusion). The curves in Figures \ref{fig:ex_partition}(b)-(d) respective to the top, bottom, and fusion channels are consistent with the assertion of Lemma \ref{fusionlem}, wherein it indicates that the partitions of the top and bottom channels are refined by the fusion channel.

5) The following DNN networks were derived by applying the DAG-closure rule R to modules.  

\begin{exam}
\begin{figure}[tp]
\centering
{\includegraphics[width=0.7\textwidth]{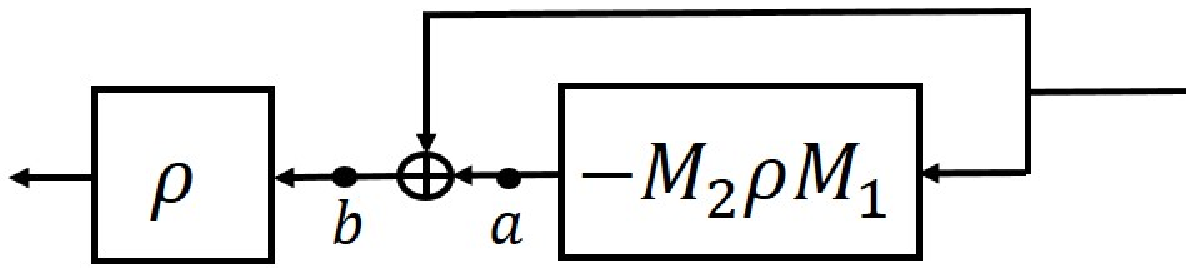}}
\caption{ResNet module featuring direct link and the same domain partitions at points $a$ and $b$. In DenseNet, the addition node is replace with the concatenation node.}
\label{figresnet}
\end{figure}
As shown in Figure \ref{figresnet}, the ResNet module \cite{he2016deep} comprises ReLU and a fusion module:
\begin{align*}
\text{(ResNet)} \doteq \rho \circ  (\text{fusion}).
\end{align*}
Using matrix notation, we obtain the following:
\begin{align} \label{resnet0}
\bx \rightarrow \rho [\bI \; \bI] \diag(\bI, - M_2 \rho M_1) \begin{bmatrix} \bI \\ \bI \end{bmatrix} \bx = \rho (\bI - M_2 \rho M_1) \bx,
\end{align}
where $M_1$ and $M_2$ are affine mappings. The unique feature of ResNet is the direct link, which enhances resistance to the gradient vanishing problem in back-propagation algorithms \cite{kingma2014adam}. The fact is that the direct linking and batch-normalization \cite{ioffe2015batch,szegedy2017inception} have become indispensable elements in the learning of very deep neural networks using back-propagation algorithms.

Let $\mathcal M_L$ denote an $L$-layer DNN. A residual network  \cite{he2016deep} extends $\mathcal M_L$ from $L$ layers to $L+2$ layers, as follows: $\text{(ResNet) } \circ  \mathcal M_L$.
Repetition of this extension allows a residual network to maintain an arbitrary number of layers. 
As noted in the caption of Figure \ref{figresnet}, domain partitioning is the same at $a$ and $b$.
This can be derived in accordance with the following analysis.
Let $\mathbb P_0$ denote the domain partitioning of $\chi$ at the input of the module. The top channel of the parallel module retains the partition, whereas in the bottom channel, the partition is refined as $\mathbb P_1$ using $M_2 \rho  M_1$. In accordance with (\ref{fusiondom}), the domain of the fusion function is $\mathbb P_1\cap \mathbb P_0$ (i.e., $\mathbb P_1$). Thus, the domain partitions at $a$ and $b$ are equivalent.  Note that the DenseNet module \cite{huang2017densely} replaces the addition node in Figure \ref{figresnet} with the concatenation node. The partitions of DenseNet at $b$ and $a$ are the same, as in the ResNet case.

\end{exam}

\begin{figure}[h]
\begin{center}
\subfigure[]{\includegraphics[width=0.9\textwidth]{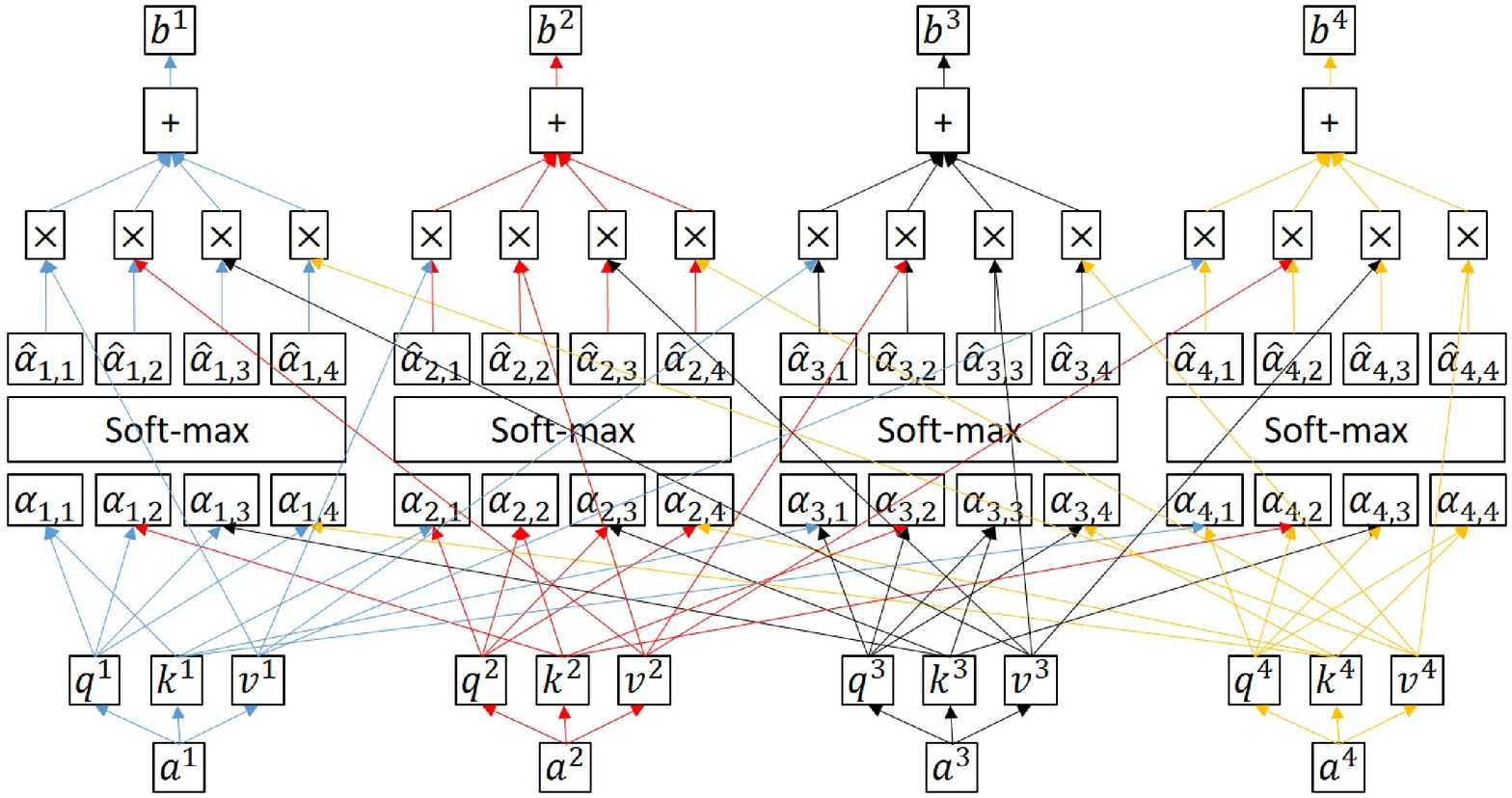}}
\subfigure[]{\includegraphics[width=0.9\textwidth]{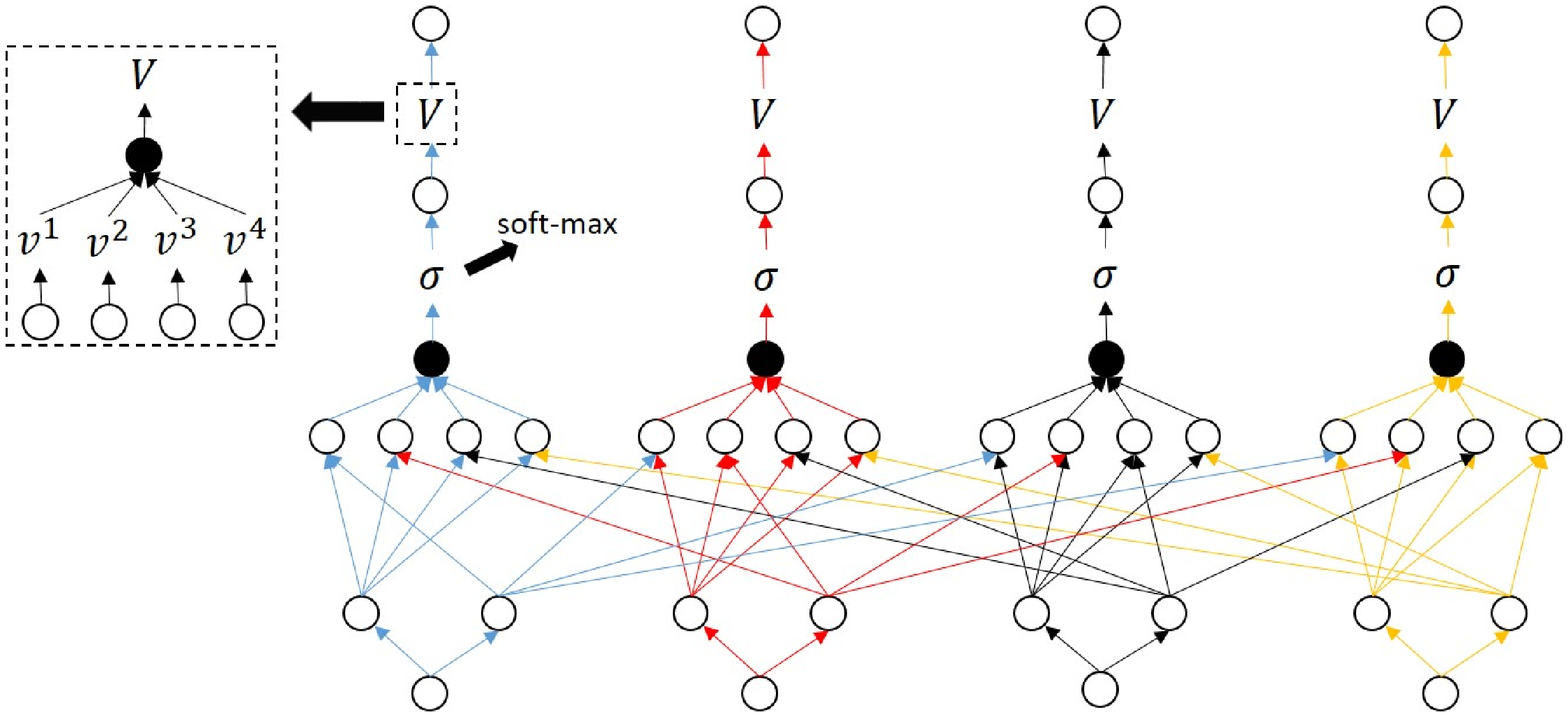}}
\end{center}\caption{Self-attention module: (a) network; (b) graph, in which the highlighted dashed-box represents the graph used to obtain dictionary $V$ from value vectors.}
\label{fig:attention}
\end{figure}

\begin{exam}
Transformers are used to deal with sequence to sequence conversions, wherein the derivation of long-term correlations between tokens in a sequence is based on the attention module \cite{vaswani2017attention,dosovitskiy2020image}. A schematic illustration of a self-attention module is presented in Figure \ref{fig:attention}(a), where the inputs vectors are $a^1$, $a^2$, $a^3$, $a^4$, and outputs vectors are $b^1$, $b^2$, $b^3$, $b^4$. The query, key, and value vectors for $a^i$ are respectively generated from matrix $W^q$, $W^k$, and $W^v$, where $q^i=W^q a^i$, $k^i=W^k a^i$, and $v^i=W^v a^i$ for all $i$. Attention score $\alpha_{i,j}$ indicates the inner product between the normalized vectors of $q^i$ and $k^j$. The vector of the attention scores $[\alpha_{i,1}$, $\alpha_{i,2}$, $\alpha_{i,3}$, $\alpha_{i,4}]^\top$ are input into the soft-max layer to obtain probability distribution $[\hat{\alpha}_{i,1}$, $\hat{\alpha}_{i,2}$, $\hat{\alpha}_{i,3}$, $\hat{\alpha}_{i,4}]^\top$, where $\hat{\alpha}_{i,j}=\frac{e^{\alpha_{i,j}}}{\sum_j e^{\alpha_{i,j}}}$. Output vector $b^i=\sum_j \hat{\alpha}_{i,j} v^j$ is derived via multiplications and additions as a linear combination of value vectors with coefficients derived from the probability distribution. Figure \ref{fig:attention} (b) presents a graphical representation of (a) wherein non-linear transformation $\sigma$ is the soft-max function. Dictionary $V$ of value vectors can be obtained by a performing concatenation operation, which implicitly involves reshaping the dimension of the resulting vector to the matrix (see  dashed-box in the figure).
\end{exam}

\begin{figure}[ht]
\begin{center}
\subfigure[]{\includegraphics[width=0.9\textwidth]{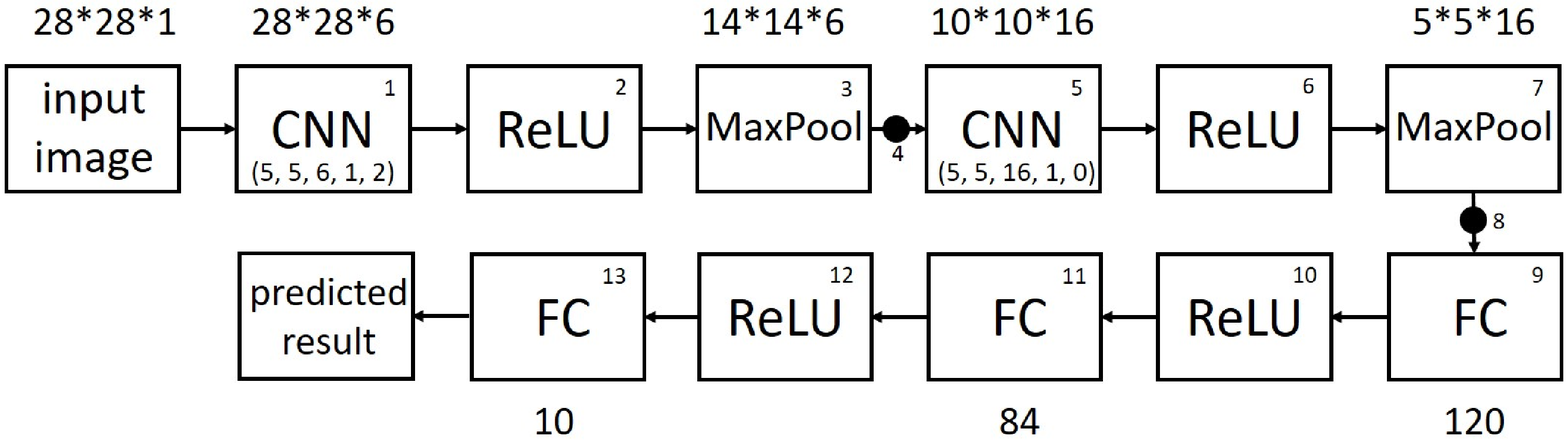}}
\subfigure[]{\includegraphics[width=0.9\textwidth]{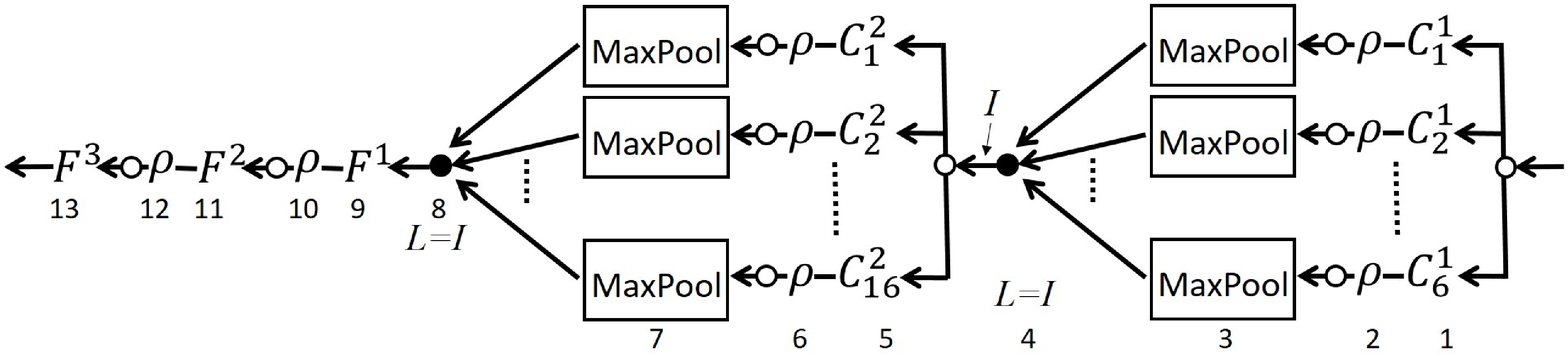}}
\subfigure[]{\includegraphics[width=0.6\textwidth]{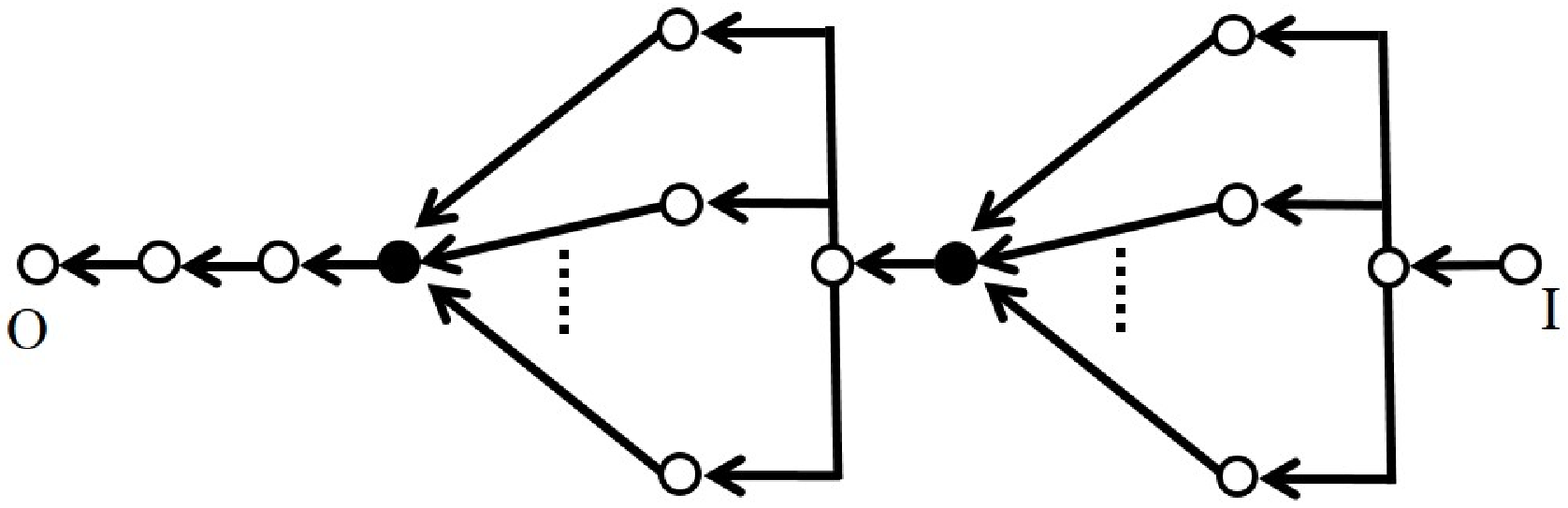}}
\end{center}\caption{LeNet-5: (a) network; (b) graph, in which the numbers beneath the nodes and arcs correspond to block numbers in (a) and solid nodes indicate fusion module/concatenation; (c) simplification of (b) illustrating composition of modules, where $I$ and $O$ respectively denote input and output. 
}
\label{fig:lenet5_ex}
\end{figure}

\begin{figure}[ht]
\begin{center}
\subfigure[]{\includegraphics[width=0.43\textwidth]{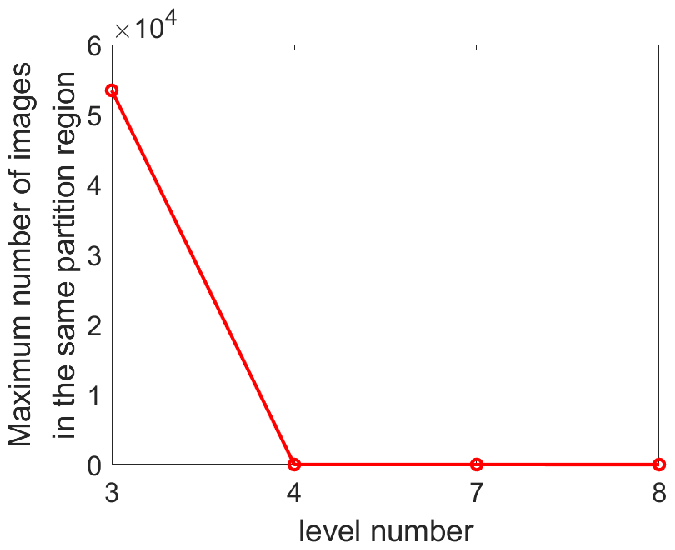}}
\subfigure[]{\includegraphics[width=0.43\textwidth]{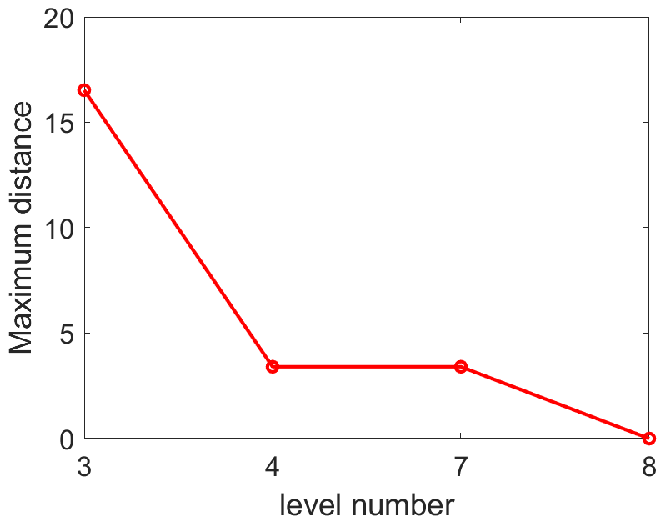}}
\subfigure[]{\includegraphics[width=0.43\textwidth]{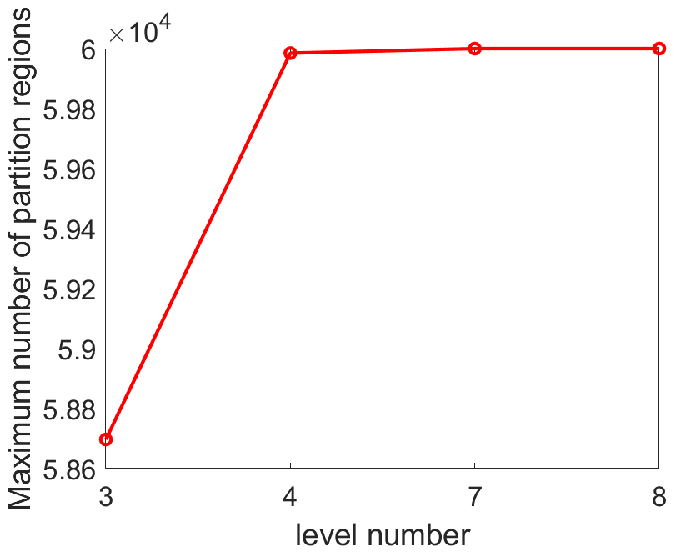}}
\end{center}\caption{Illustrations of partition refinements in LeNet-5 at output of levels $3$, $4$, $7$, and $8$ (as shown in Figure \ref{fig:lenet5_ex}(b)) where input is $60,000$ images from MNIST dataset (note that levels $3$ and $7$ are parallel connections linking several channels; therefore, only the maximum values of all channels at those levels are plotted): (a) maximum number of images in a given same partition region; (b) maximum distance between pairs of images located in the same partition region (distance at level $8$ is zero, which means that each partitioning region contains no more than one image); (c) maximum number of partition regions in a channel (note that as one moves up to the next levels, the curves in (a) and (b) decrease, while the curve in (c) increases, which is consistent with our assertion that the partitions of the previous channels are refined by the fusion channel).}
\label{fig:lenet5_partition}
\end{figure}

\begin{exam}
Figure \ref{fig:lenet5_ex} illustrates the well-known LeNet-5 network \cite{lecun1998gradient}. Figure \ref{fig:lenet5_ex}(a) presents a block diagram of the network, in which the input is an image ($28\times 28$ px) and the output is ten classes of characters ($0$ to $9$). CNN block number $1$ is a convolution layer with the following parameters: filter size ($5\times5$), stride ($1$), padding ($2$), and $6$-channel output (six $28\times 28$ images). Block numbers $2$ and $3$ indicate MaxLU operations and $6$-channel output (six $14\times 14$ images). Black circle $4$ indicates a concatenation operation (O2), the output of which is an image. Block number $5$ indicates a convolution layer with the following parameters: filter size ($5\times 5$), stride ($1$), and padding ($0$). This layer outputs sixteen $10\times 10$ images. Block numbers $6$ and $7$ indicate MaxLU operations,  the output of which is sixteen $5\times 5$ images. Black circle $8$ indicates a concatenation operation (O2) and the output is a vector. Block number $9$ indicates a fully-connected network with input dimensions of $400$ and output dimensions of $120$. Block number $10$ indicates a ReLU activation function. Block number $11$ is a fully-connected network with the input dimensions of $120$ and output dimensions of $84$. Block number $12$ indicates a ReLU activation function. Block number $13$ indicates a fully-connected network with input dimensions of $84$, where the output is a prediction that includes $1$ of the $10$ classes. Figure \ref{fig:lenet5_ex}(b) presents a graphical representation of (a), and (c) presents a simplified graphical representation of (a). In Figure \ref{fig:lenet5_ex}(c), we can see that LeNet-5 begins with a sequence of compositions of modules (featuring a parallel module followed by a fusion module), and then a sequence of MaxLU layers.
Figure \ref{fig:lenet5_partition} illustrates properties of  partitions at the outputs of levels $3$, $4$, $7$, and $8$ in Figure \ref{fig:lenet5_ex}(b). The curves in Figures \ref{fig:lenet5_partition}(a)-(c) are consistent with the assertion of Lemma \ref{fusionlem}, which indicates that the partitions of the previous channels are refined by the fusion channel.
\end{exam}

\section{Properties of general DNNs} \label{commonprop}

In accordance with the axiomatic approach, we define general DNNs (i.e., $\cK$) representable using DAGs. In the following, we outline the properties belonging to all members in the class.

\subsection{Function approximation via partition refinement}

\begin{figure}[tp]
\centering
\subfigure[]{\includegraphics[width=0.8\textwidth]{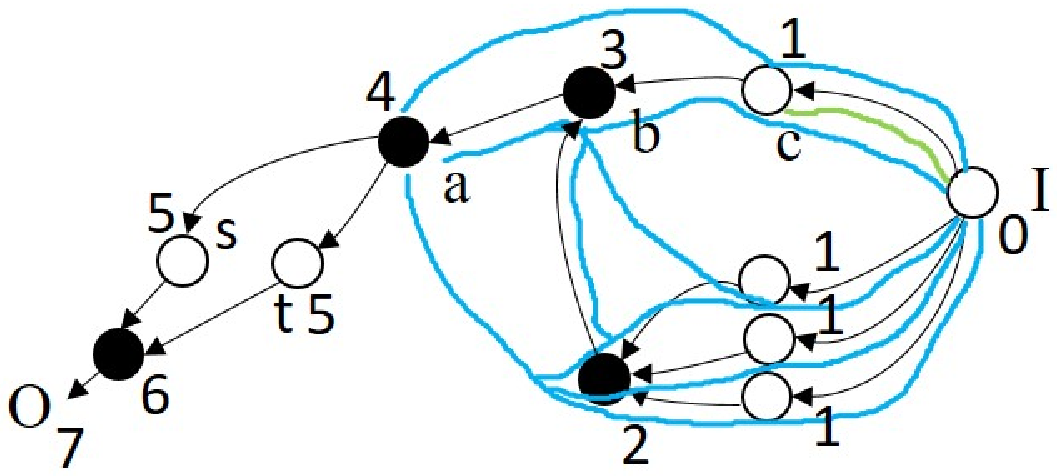}}
\subfigure[]{\includegraphics[width=0.35\textwidth]{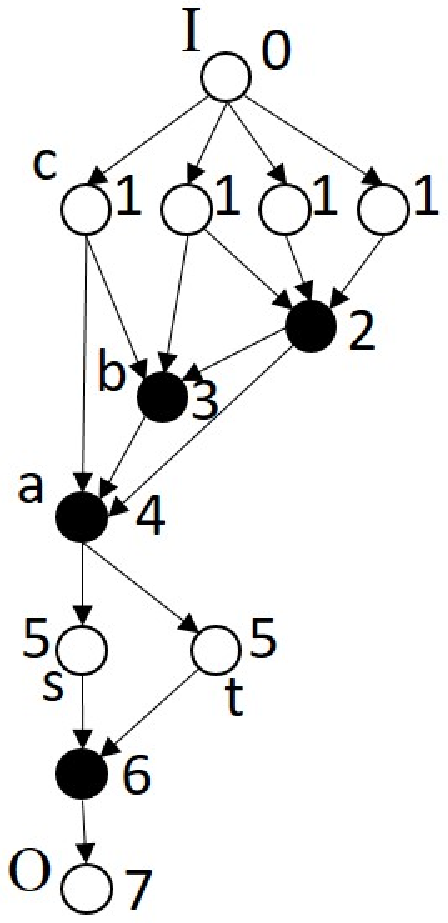}}
\caption{DAG representation of DNN illustrating the computable sub-graphs and levels of nodes (solid nodes denote fusion/concatenation nodes and $I$ and $O$ denote input and output, respectively): (a) Computable sub-graph associated with each node. Light blue denotes the computable sub-graph of node $a$, wherein the longest path from input node to node $a$ contains four arcs (i.e., $l(a) = 4$). Light green denotes the computable sub-graph of $c$, wherein the longest path from input node to node $c$ contains one arc (i.e., $l(c) = 1$); (b) Nodes in (a) are partially ordered in accordance with levels (e.g., level $1$ has four nodes; level $2$ has one node).
}
\label{figgraph}
\end{figure}

We define the computable sub-graph of node $a$ as the sub-graph of the DAG containing precisely all the paths from the input node of the DAG to node $a$.  Clearly, the computable sub-graph defines a DNN in $\cK$, the output node node of which is $a$, such that it computes a CPWL function. 
In Figure \ref{figgraph}(a), the computable sub-graph of node $a$ (highlighted in light blue) contains nodes with numbers $0$, $1$, $2$, $3$, and $4$.  The computable sub-graph of node $c$ (highlighted in light green) is contained in the sub-graph of $a$. 

In the following, we outline the domain refinement of a general DNN. Specifically, if node $b$ is contained in a computable sub-graph of $a$, then the domain partition imposed by that sub-graph is a refinement of the partition imposed by computable sub-graph of $b$.

\begin{citedthm}\label{refinement}
Let the domain partitions imposed by computable sub-graphs $\mathcal G^a$ (at node $a$) and $\mathcal G^b$ (at node $b$) of a general DNN be respectively $\mathbb P_a$ and $\mathbb P_b$. Suppose further that node $b$ is contained in sub-graph $\mathcal G^a$. Then $\mathbb P_a$ refines $\mathbb P_b$.
\end{citedthm}
\proof
Without a loss of generality, we suppose that arcs in the general DNN are atomic operations and the functions applied to arcs are in base set $\cB$.
Suppose further that $p$ is a path from the input node to node $a$, which also passes through node $b$. Let $p'$ denote the sub-path of $p$ from node $b$ to node $a$ ($p':  b=c_0 \rightarrow c_1 \cdots \rightarrow a=c_n$). 
Let $\mathbb P_{c_i}$ denote the partition of input space $\chi$ defined using the computable sub-graph at node $c_i$. In the following, we demonstrate that if $c_i \rightarrow c_{i+1}$, then $\mathbb P_{c_i}$ is refined by $\mathbb P_{c_{i+1}}$.  Note that arc $c_i \rightarrow c_{i+1}$ belongs to one of the three atomic operations. If it is a series-connection operation (O1), then the refinement is obtained by referring to Lemma \ref{seriesprop}. If it is a concatenation operation (O2), then the refinement is obtained by referring to Lemma \ref{fusionlem}; If it is a duplication operation (O3), then the partitions for nodes $c_i$ and $c_{i+1}$ are the same. Thus, $\mathbb P_a$ is a refinement of $\mathbb P_b$.

\qed

Figure \ref{figgraph}(a) presents the DAG representation of a DNN. Node $b$ is contained in the computable sub-graph of node $a$, whereas node $c$ is contained in the computable sub-graph of $b$, such that the domain partition of $a$ is a refinement of the partition of $b$ and the domain partition of $b$ is a refinement of the partition of $c$. Thus, the domain partition of node $a$ is a refinement of the domain partition of node $c$.

\begin{figure}[th]
\begin{center}
\subfigure{\includegraphics[width=0.4\textwidth]{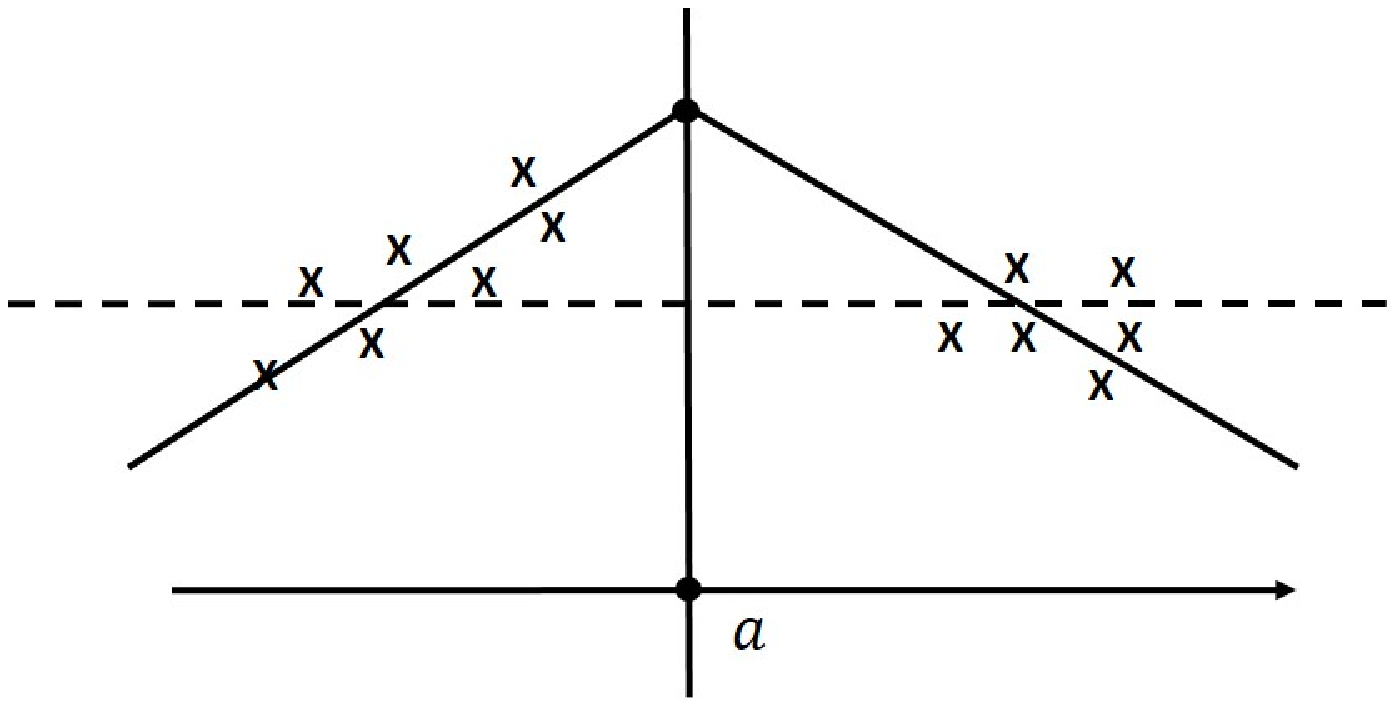}}
\end{center}
\caption{Approaches to linear regression: The dashed line indicates the regression line derived from all data (i.e., conventional approach); and the solid lines indicate regression lines derived from data of $x > a$ and data of $x \leq a$  continuous at $x=a$ (i.e., general DNN approach).} \label{figregression}
\end{figure}

As hinted in Theorem \ref{refinement}, general DNNs are implemented using a  data-driven ``divide and conquer" strategy when performing function approximation. In other words, when travelling a path from the input node to a node of the DNN, we can envision the progressive refinement of the input space partition along the path, where each partition region is associated with an affine linear mapping.  Thus, computing a function using a general DNN corresponds to approximating the function using local simple mappings over regions in a partition derived using the DNN. A finer approximation of the function can be obtained by increasing the lengths of paths from the input node to the output node. 
Figure \ref{figregression} illustrates the conventional and general DNN approaches to the problem of linear regression. The conventional approach involves fitting ``all" of the data to obtain a dashed hyperplane, whereas the general DNN approach involves dividing the input space into two parts and fitting each part using hyperplanes.


\subsection{Stability via sparse/compressible weight coefficients}

We introduce the $l$-function of node $a$ to denote the number of arcs along the longest path from the input node $I$ to node $a$ of a DAG. We refer to $l(a)$ as the level of node $a$. According to this definition, the level of the input node is zero. 
\begin{citedlem} \label{l-function}
The level is a continuous integer over the nodes in a general DNN. In other words, if node $a$ is not the output node, then there must exist a node $b$ where $l(b) = l(a) + 1$. 
\end{citedlem}

\proof
Clearly, $l(I) = 0$ for input node $I$.
This lemma can be proven via contradiction. Suppose that the levels are non-continuous integers. Without a loss of generality, the nodes can be divided into two groups  ($A$ and $B$), where $A$ includes all of the nodes with level $\leq n$ and $B$ includes all of the nodes with level $\geq n+k$ and $k > 1$.  Let $b \in B$ have the smallest level in graph $B$ and $l(b) = n+k$. Further, let $p(b)$ denote a longest path from the input node to node $b$ and let $a$ be a node along the path with arc $a \rightarrow b$. Thus, $l(a) < n+k$; otherwise, $l(b) \geq n+k+1$, which violates the assumption that $l(b) = n+k$. If $l(a) \leq n$, then $l(b) \leq  n+1$ (since $a$ is on the longest path $p(b)$ to $b$ and $a$ has a direct link to $b$). This violates the assumption $l(b) = n+k$ with $k > 1$. Thus, $n < l(a) < n +k$ from which $a \notin A$ and $a \notin B$. This violates the assumption that all nodes can be divided into two groups  ($A$ and $B$). We obtain a contradiction and hence complete the proof.
\qed

The nodes in a DAG can be ordered in accordance with the levels.  Assume that there is only one output node, denoted as $O$. Clearly, $l(O) = L$ is the largest level associated with that DAG. The above lemma implies that the nodes can be partitioned into levels from $0$ to $L$. We introduce notation $\bar l(n)$ (referring to the nodes at level $n$) to denote the collection of nodes with levels equal to $n$; and let $| \bar l(n)|$ denote the number of nodes at that level.  As shown in Figure \ref{figgraph} (b), the nodes in Figure \ref{figgraph}(a) are ordered in accordance with their levels, as indicated by the number besides the nodes.  For any DNN $\mathcal N_L \in \cK$, we can define DNN function $\mathcal N_n$ (with $n \leq L$) by stacking the DNN functions of nodes at level $n$ into a vector, as follows:
\begin{align} \label{leveln1}
\mathcal N_n= [\mathcal N_{(a)}]_{a \in \bar l(n)}
\end{align}
where 
$N_{(a)}$ is the function derived using the computable sub-graph of node $a \in \bar l(n)$. Clearly, $\mathcal N_n \in \cK$ because it is formed by concatenation of $\mathcal N_{(a)}$. For example, in Figure \ref{figgraph}(b), $\mathcal N_3 = [ \mathcal N_{(b)}]$ and $\mathcal N_5 = \begin{bmatrix} \mathcal N_{(s)} \\ \mathcal N_{(t)} \end{bmatrix}$. The order of components in $N_n$ is irrelevant to sequent analysis of stability conditions.

The stability of a DNN can be measured as the output perturbation against the input perturbation, such that  
\begin{align} \label{DAGstable}
\| \mathcal N_L (\bx)- \mathcal N_L(\by) \|_2\leq C(L) \| \bx - \by\|_2,
\end{align}
where $L$ is the level of the output node for DNN $\mathcal N_L$. A stable deep architecture implies that a deep forward inference is well-posed and robust in noisy environments.
A sufficient condition for the DNN to be stable requires that $C(L)$ be a bounded non-increasing function of $L$ when $L \rightarrow \infty$. 

\begin{citedlem} \label{LocalLip}
Let $d$ be the uniform bound defined in (\ref{uniformbound}); let $I_a = \{b | b \rightarrow a\}$ (the nodes directly linking to the node $a$); and let $|I_a|$ denote the number of nodes in $I_a$. Further, let $\mathcal N_L \in \cK$ and let $\mathcal N_{L, \textit p}$ denote the restriction of $\mathcal N_L$ over partition region $\textit p$. Suppose that $ \mathcal N_{(a)}$ and $ \mathcal N_{(a),p}$ respectively denote the CPWL function associated with computable sub-graph of node $a$ and the restriction of the function over $\textit p$. Further, as defined in (\ref{leveln1}), $\mathcal N_{n}$ denotes the function derived by 
nodes at level $n \leq L$
and $\mathcal N_{n, \textit p}$ denotes the restriction of $\mathcal N_{n}$ on 
domain partition ${\textit p}$; i.e., 
\begin{align}\label{levelnp1}
\mathcal N_{n, \textit p} = [ \mathcal N_{(a), \textit p}]_{a \in \bar l(n)}.
\end{align}
(i) For given $n$ and $\textit p$, there exists $C(n, p)$, such that for any $\bx, \by \in \textit p$,
\begin{align} \label{DAGstable1}
\| \mathcal N_{n, \textit p} (\bx)- \mathcal N_{n, \textit p}(\by) \|_2\leq C(n, \textit p) \| \bx - \by\|_2,
\end{align}
where $C(n, \textit p)$ is referred to as the Lipschitz constant in $\textit p$ at level $n$. \\
(ii) If there exists level $m$, such that for $n  \geq  m$, 
\begin{align} \label{stab3}
d\sum_{a \in \bar l(n)} \sum_{b \in I_a}   \| \bW_{ab}\|_2  \leq 1,
\end{align}
where $\bW_{ab}$ is the weight matrix associated with the atomic operation on arc $a \leftarrow b$, then the Lipschitz constant $C(n, \textit p)$ is a bounded function of $n$ on $\textit p$.  
\end{citedlem}
\proof 
See Appendix~\ref{appendixa} for the proof.
\qed

This lemma establishes local stability in a partition region of a DNN. To achieve global stability in the input space, we invoke the lemma in \cite{Wen19}, which indicates that piece-wise functions persisting local stability have global stability, provided that the functions are piece-wise continuous.

\begin{lem}\cite{Wen19} \label{GlobalLip}
Let $\chi = \cup_i P_i $ be a partition and $f_i$ with domain $\overline{P_i}$ be $l_i$-Lipschitz continuous with $f_i(\bx) = f_j(\bx)$ for $\bx \in\overline{P_i} \cap\overline{P_j}$. Let $f$ be defined by $f(\bx) := f_i(\bx)$ for $\bx \in P_i$. Then, $f$ is $(\max_i l_i)$-Lipschitz continuous.
\end{lem}

\begin{citedthm}\label{thmstabsuff}
For stability, we hold the assumption pertaining to Lemma \ref{LocalLip}. Let DNN $\mathcal N_L \in \cK$ with the domain on input space $\chi$ and let $\mathcal N_{n}$ denote the function with nodes of $\mathcal N_L$ up to level $n \leq L$.
If there exists level $m$ such that for $n \geq  m$,
\begin{align} \label{stabsuff}
d\sum_{a \in \bar l(n)} \sum_{b \in I_a}   \| \bW_{ab}\|_2  \leq 1
\end{align}
where $d$ is the uniform bound defined in (\ref{uniformbound}), $\bW_{ab}$ is the weight matrix associated with arc $a \leftarrow b$, then $C(L) = \max_{\textit p} C(L, \textit p)$, where $\textit p$ is a partition region of $\chi$, is a bounded non-increasing function of $L$ on $\chi$. Note that $C(L)$ is defined in (\ref{DAGstable}). Hence, $\mathcal N_L$ is a stable architecture.  
\end{citedthm}
\proof 
See Appendix~\ref{appendixb} for the proof.
\qed

\begin{figure}[th]
\centering
{\includegraphics[width=0.4\columnwidth]{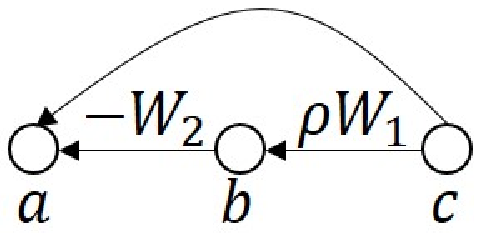}}
\caption{Conditions for stability of
ResNet module $\bI -  \bW_2 \text{(ReLU)} \bW_1$ where $I_a = \{b, c\}$: If (\ref{stabsuff}) is satisfied, then $\| \bW_2\|_2 = 0$ (hence, $\bW = 0$) due to the fact that  
$1 + \|\bW_2\|_2 \leq 1$ if and only if $\|\bW_2\|_2 = 0$. In fact, it is sufficient for the model to achieve stability if  $\bW \neq 0$ with $\|\bI - \bW_2 \bW_1\|_2 \leq 1$ (all eigenvalues of $\bW_2 \bW_1$ are lying in $[0, 1]$).} \label{comResNet}
\end{figure}

This theorem extends the stability of the series-connected DNNs in \cite{Wen19} to general DNNs. 
From $\| \bW\|_F \geq \| \bW\|_2$, the condition determining the stability of an DNN can be expressed as follows:
\begin{align} \label{stabsuff1}
d\sum_{a \in \bar l(n)} \sum_{b \in I_a}   \| \bW_{ab}\|_F  \leq 1.
\end{align}
This condition holds, regardless of the size of $\bW_{ab}$. Thus, if the matrix is large, then (\ref{stabsuff1}) implies that the weight coefficients are sparse/compressible.
Note that (\ref{stabsuff}) is a sufficient condition for a DNN to achieve stability, however, it is not a necessary condition. The example in Figure \ref{comResNet} demonstrates that satisfying (\ref{stabsuff}) is not a necessary.

\begin{figure}[!h]
\begin{center}
  \subfigure[]{\includegraphics[width=0.45\textwidth]{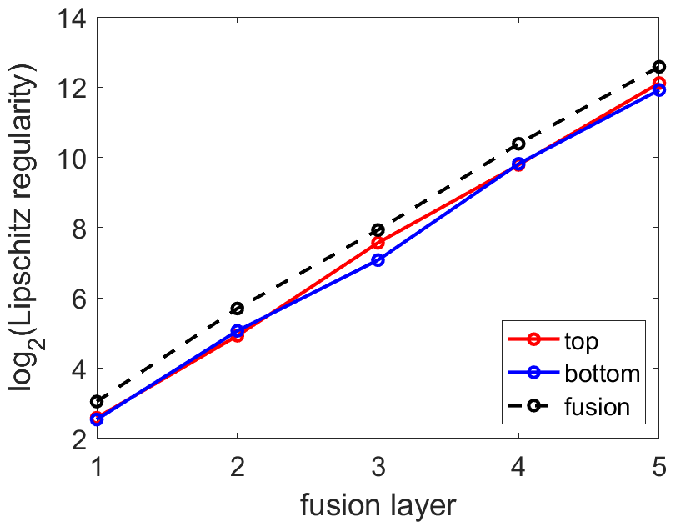}}
  \subfigure[]{\includegraphics[width=0.45\textwidth]{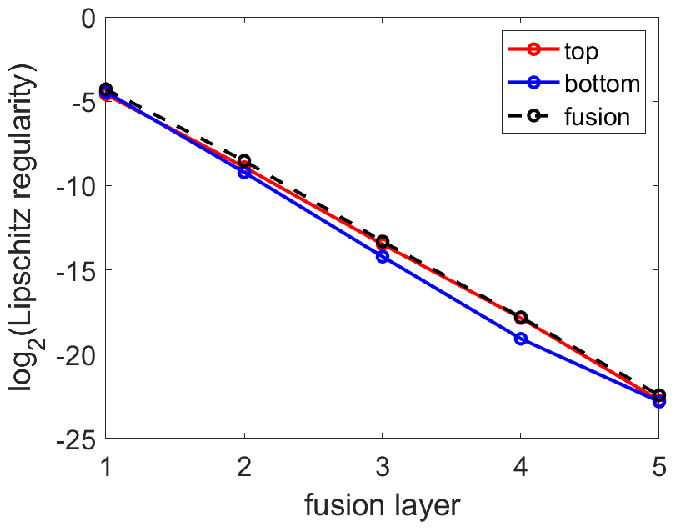}}
\end{center}\caption{Stability simulation using the network in Figure \ref{fig:ex_partition}(a), in which the   maximum gain $\|\mathcal N_L\bx-\mathcal N_L\by\|_2/\|\bx-\by\|_2$ for all training pairs is plotted against each fusion layer $j$, where $j=1, \cdots, 5$
(dimensions of the weight matrices and biases in $M_i$ in top, bottom, and fusion channels of each module are respectively $20\times20$ and $20\times1$;  coefficients in $M_i$ are sampled i.i.d. from standard normal distribution (mean zero and variance of one); inputs are $2,000$ random vectors (each of size $20\times1$) with entries sampled i.i.d. from the standard normal distribution): (a) Maximum gain increases with an increase in the number of the fusion layers; (b) Maximum gain remains bounded when weight coefficients in $M_i$ are scaled to meet  (\ref{stabsuff1}) for Theorem \ref{thmstabsuff}. Note that $d = 1$ for ReLU activation functions.}
\label{fig:ex_lipschitz1}
\end{figure}

Figure \ref{fig:ex_lipschitz1} illustrates simulations pertaining to Theorem \ref{thmstabsuff} for the network presented in Figure \ref{fig:ex_partition}(a). In Figure \ref{fig:ex_lipschitz1}(a), the upper bound for Lipschitz constant $C(L)$ where $L=1, \cdots, 5$ (presented as the maximum gain for all pairs of training data) increases with an increase in the fusion layers if the weight coefficients do not satisfy (\ref{stabsuff}) in terms of stability. If the weight coefficients were scaled to be compressible in accordance with (\ref{stabsuff}), then the upper bound would decrease with an increase in the fusion layers, as shown in Figure \ref{fig:ex_lipschitz1}(b). This is an indication that the network is stable versus the input perturbation.

\section{Conclusions} \label{conclusions}

Using an axiomatic approach, we established a systematic approach to representing deep feedforward neural networks (DNNs) as directed acyclic graphs (DAGs). The class of DNNs constructed based on the approach is referred to as general DNNs, which covers pragmatic modules, activation functions, and non-linear transformations in DNNs. 
Un-rectifying analysis revealed function approximation and stability properties to general DNNs. 
We demonstrate that general DNNs approximate a (known/unknown) function via data learning in a coarse-to-fine manner through the refinement of input space partitions. A partition can be refined using a composition of activation functions or a fusion operation combining inputs from more than one path to a node. If the weight coefficients on arcs become increasing sparse along any path of a graph, then the DNN function gains stability with respect to perturbations in the input space, due to a bounded global Lipschitz constant for the network. These properties imply that a general DNN ``divides" the input space, ``conquers" each partition with simple approximating function, and ``sparsifies" weight coefficients to gain robustness against input perturbations. Implementing the axiomatic approach (i.e., the atomic operations, basic elements, and regulatory rules) in conjunction with the un-rectifying of activation functions makes it possible to generate graph representations that can be used for the analysis of DNNs provided that the host of graph analysis tools can be suitably leveraged to study the structure of DNNs.

\noindent{\bf{Acknowledgements}}: An error in the original proof of Lemma \ref{LocalLip} was corrected by Mr. Ming-Yu Chung.

\bibliographystyle{ieeetr}
\bibliography{DAG}

\appendix

\section{Proof of Lemma \ref{LocalLip}} \label{appendixa}

\proof
(i) Without a loss of generality, the arcs in $\mathcal N_L$ are atomic operations associated with functions in basis set $\cB$. 
The base step is on input node $I$ (i.e., $\bar l(0)$). Clearly, (\ref{DAGstable1}) holds when $C(0, \textit p) =1$.

Implementation of the induction step is based on levels. Suppose that (\ref{DAGstable1}) holds for all nodes in levels smaller than $n$. If $a$ is a node in level $n$, then node $b \in I_a$ must be in level $l(b) <  n$. Thus, for $\bx \in \textit p$, 
\begin{align} 
\mathcal N_{(a), \textit p}
= 
[B_{ab} \mathcal N_{(b), \textit p}(\bx)]_{b \in I_{a}},
\end{align}
where $B_{ab} \in \cB$, in accordance with the fact that functions associated with axiomatic operations on arcs are members in $\cB$. If the atomic operation is a duplication, then $B_{ab} = \bI$;  and if the atomic operation is a series-connection/concatenation, then $B_{ab} \in \{\bI,\rho \bL, M, \rho M, \sigma \bL, \sigma M\}$. Further,
\begin{align} 
\|\mathcal N_{(a), \textit p}(\bx) - \mathcal N_{(a), \textit p}(\by)\|_{2}
=
\sum_{b \in I_a} \|B_{ab} (\mathcal N_{(b), \textit p}(\bx)- \mathcal N_{(b), \textit p}(\by))\|_{2} .
\end{align}
Case 1. Consider $B_{ab} \in \{\bI,\rho \bL, M, \rho M\}$. For $\bx, \by \in \textit p$, the bias term in $B_{ab}$ (if any) can be cancelled. 
Applying the uniform bound assumption on activation functions (\ref{uniformboundact}) and applying (\ref{levelnp1}) and  (\ref{DAGstable1}) to levels smaller than $n$ results in the following: 
\begin{align} \label{upbd}
\| \mathcal N_{n, \textit p}(\bx) - \mathcal N_{n, \textit p}(\by) \|_2
& \leq 
\sum_{a \in \bar l(n)}\sum_{b \in I_a} d_{\rho} \|\bW_{ab} \|_2 \|\mathcal N_{(b), \textit p}(\bx) - \mathcal N_{(b), \textit p}(\by)\|_2  \nonumber \\
& \leq  
\sum_{a \in \bar l(n)}\sum_{b \in I_a}  d_{\rho} \| \bW_{ab}\|_2 C(l(b), \textit p)  \|\bx - \by\|_2 . 
\end{align}

Case 2. Consider $B_{ab} \in \{\bI,\sigma \bL, M, \sigma M\}$. Similarly, for $\bx, \by \in \textit p$, we can obtain following:
\begin{align} \label{upbdsoft}
\| \mathcal N_{n, \textit p}(\bx) - \mathcal N_{n, \textit p}(\by) \|_2
& =
\sum_{a \in \bar l(n)}\sum_{b \in I_a}  \|\sigma( M_{ab}\circ\mathcal N_{(b), \textit p}(\bx)) - \sigma( M_{ab}\circ\mathcal N_{(b), \textit p}(\by))\|_2  \nonumber \\
& \leq
\sum_{a \in \bar l(n)}\sum_{b \in I_a}  d_{\sigma}\|M_{ab}\circ\mathcal N_{(b), \textit p}(\bx) -  M_{ab}\circ\mathcal N_{(b), \textit p}(\by)\|_2  \nonumber \\
& \leq 
\sum_{a \in \bar l(n)}\sum_{b \in I_a} d_{\sigma} \|\bW_{ab} \|_2 \|\mathcal N_{(b), \textit p}(\bx) - \mathcal N_{(b), \textit p}(\by)\|_2  \nonumber \\
& \leq  
\sum_{a \in \bar l(n)}\sum_{b \in I_a} d_{\sigma} \| \bW_{ab}\|_2 C(l(b), \textit p)  \|\bx - \by\|_2 ,
\end{align}
where $\bW_{ab}$ is the linear part of the affine function $M_{ab}$ and $d_{\sigma}$ is the Lipschitz constant bound defined in (\ref{uniformboundfor}) .
Finally, $(\ref{upbd})$ and $(\ref{upbdsoft})$ are combined using (\ref{uniformbound}) to yield
\begin{align} \label{stab2}
C(n, \textit p)  = d \sum_{a \in \bar l(n)}\sum_{b \in I_a}  \| \bW_{ab}\|_2 C(l(b), \textit p),
\end{align}
where $l(b) \leq n-1$.
This concludes the proof of (i). \\

(ii) 
Let $C^*(m-1, \textit p)$ denote the maximal value of $C(k, \textit p)$ for $k \leq m-1$.
From (\ref{stab2}) and (\ref{stab3}), we obtain the following:
\begin{align}\label{DAGstable2}
C(m,\textit p) \leq  d \sum_{a \in \bar l(m)}\sum_{b \in I_a}  \| \bW_{ab}\|_2 C^*(m-1, \textit p) \leq C^*(m-1, \textit p).
\end{align}

Hence, $C^*(m, \textit p) = C^*(m-1, \textit p)$. Considering the fact that (\ref{stab3}) holds for $n \geq m$, we  obtain
\begin{align} \label{stab5}
C(n,\textit p) \leq  d \sum_{a \in \bar l(n)}\sum_{b \in I_a}  \| \bW_{ab}\|_2 C^*(n-1, \textit p) \leq C^*(n-1, \textit p) 
\end{align}
Thus, $C^*(n, \textit p) = C^*(n-1, \textit p)$. Based on (\ref{DAGstable2}) and (\ref{stab5}), we obtain
\begin{align} \label{DAGstable3}
C^*(n, \textit p) = C^*(m-1, \textit p), \quad \forall n\geq m.
\end{align}
The fact that $C(n, \textit p) \leq C^*(n, \textit p)$ leads to the conclusion that $C(n, \textit p)$ is a bounded sequence of $n$ on $\textit p$.

\qed

\section{Theorem \ref{thmstabsuff}} \label{appendixb}

\proof
In accordance with (\ref{stabsuff}) and Lemma \ref{LocalLip}, $C(n, p)$ is bounded for any $p$ and $n$.
The fact that activation functions of $\mathcal N_L$ satisfy (A1) implies that the total number of partitions induced by an activation function is finite. Thus, the number of partition regions induced by $\mathcal N_n$ is finite. Hence, 
\begin{align*}
C(n) = \max_p C(n, p) 
\end{align*}
is defined and bounded above, for any $n$. In accordance with Lemma \ref{GlobalLip} and the definition of $C(n)$, 
we obtain
\begin{align}
\|\mathcal N_n(\bx) - \mathcal N_n(\by)\|_2 \leq C(n) \| \bx - \by\|_2.
\end{align}
We define $C^*(m) = \max_{k \leq m}\max_{\textit p} C(k, \textit p)$ and obtain for any $L \geq n \geq m$
\begin{align} \label{DAGstable4}
 C(n) \leq C^*(m).
\end{align}
The sequence $\{C(n)\}_{n \geq m}^L$ is bounded for any $L$, such that $\mathcal N_L$ is stable as $L \rightarrow \infty$.

\qed

\end{document}